\documentclass[lettersize,journal]{IEEEtran}
\usepackage{amsmath,amsfonts}
\usepackage{algorithmic}
\usepackage{algorithm}
\usepackage{array}
\usepackage[caption=false,font=normalsize,labelfont=sf,textfont=sf]{subfig}
\usepackage{textcomp}
\usepackage{stfloats}
\usepackage{url}
\usepackage{verbatim}
\usepackage{graphicx}
\usepackage{cite}
\usepackage{booktabs}
\usepackage{hyperref} %
\begin{document}

\title{DreamJourney: Perpetual View Generation with \\ Video Diffusion Models}

\author{Bo Pan, Yang Chen, Yingwei Pan, Ting Yao,~\IEEEmembership{Senior Member,~IEEE}, Wei Chen, and Tao Mei,~\IEEEmembership{Fellow,~IEEE}
\thanks{Bo Pan and Wei Chen are with the State Key Lab of CAD\&CG, Zhejiang University, and
Wei Chen is also with the Laboratory of Art and Archaeology Image. E-mail: bopan@zju.edu.cn; chenvis@zju.edu.cn. }
\thanks{Yang Chen, Yingwei Pan, Ting Yao, and Tao Mei are with HiDream.ai. E-mail: c1enyang.ustc@gmail.com; panyw.ustc@gmail.com; tingyao.ustc@gmail.com; tmei@live.com.}
\thanks{Corresponding author: Wei Chen and Yang Chen.}}

\markboth{Journal of \LaTeX\ Class Files,~Vol.~14, No.~8, August~2021}%
{Shell \MakeLowercase{\textit{et al.}}: A Sample Article Using IEEEtran.cls for IEEE Journals}


\maketitle

\begin{abstract}
Perpetual view generation aims to synthesize a long-term video corresponding to an arbitrary camera trajectory solely from a single input image. Recent methods commonly utilize a pre-trained text-to-image diffusion model to synthesize new content of previously unseen regions along camera movement. However, the underlying 2D diffusion model lacks 3D awareness and results in distorted artifacts. Moreover, they are limited to generating views of static 3D scenes, neglecting to capture object movements within the dynamic 4D world. To alleviate these issues, we present DreamJourney, a two-stage framework that leverages the world simulation capacity of video diffusion models to trigger a new perpetual scene view generation task with both camera movements and object dynamics. Specifically, in stage I, DreamJourney first lifts the input image to 3D point cloud and renders a sequence of partial images from a specific camera trajectory. A video diffusion model is then utilized as generative prior to complete the missing regions and enhance visual coherence across the sequence, producing a cross-view consistent video adheres to the 3D scene and camera trajectory. Meanwhile, we introduce two simple yet effective strategies (early stopping and view padding) to further stabilize the generation process and improve visual quality. Next, in stage II, DreamJourney leverages a multimodal large language model to produce a text prompt describing object movements in current view, and uses video diffusion model to animate current view with object movements. Stage I and II are repeated recurrently, enabling perpetual dynamic scene view generation. Extensive experiments demonstrate the superiority of our DreamJourney over state-of-the-art methods both quantitatively and qualitatively. Our project page: \url{https://dream-journey.vercel.app/}.
\end{abstract}

\begin{IEEEkeywords}
Perpetual View Generation, Video Diffusion Model, Multimodal Large Language Model.
\end{IEEEkeywords}

\section{Introduction}
\begin{figure*}[t]
\centering
\includegraphics[width=0.95\textwidth]{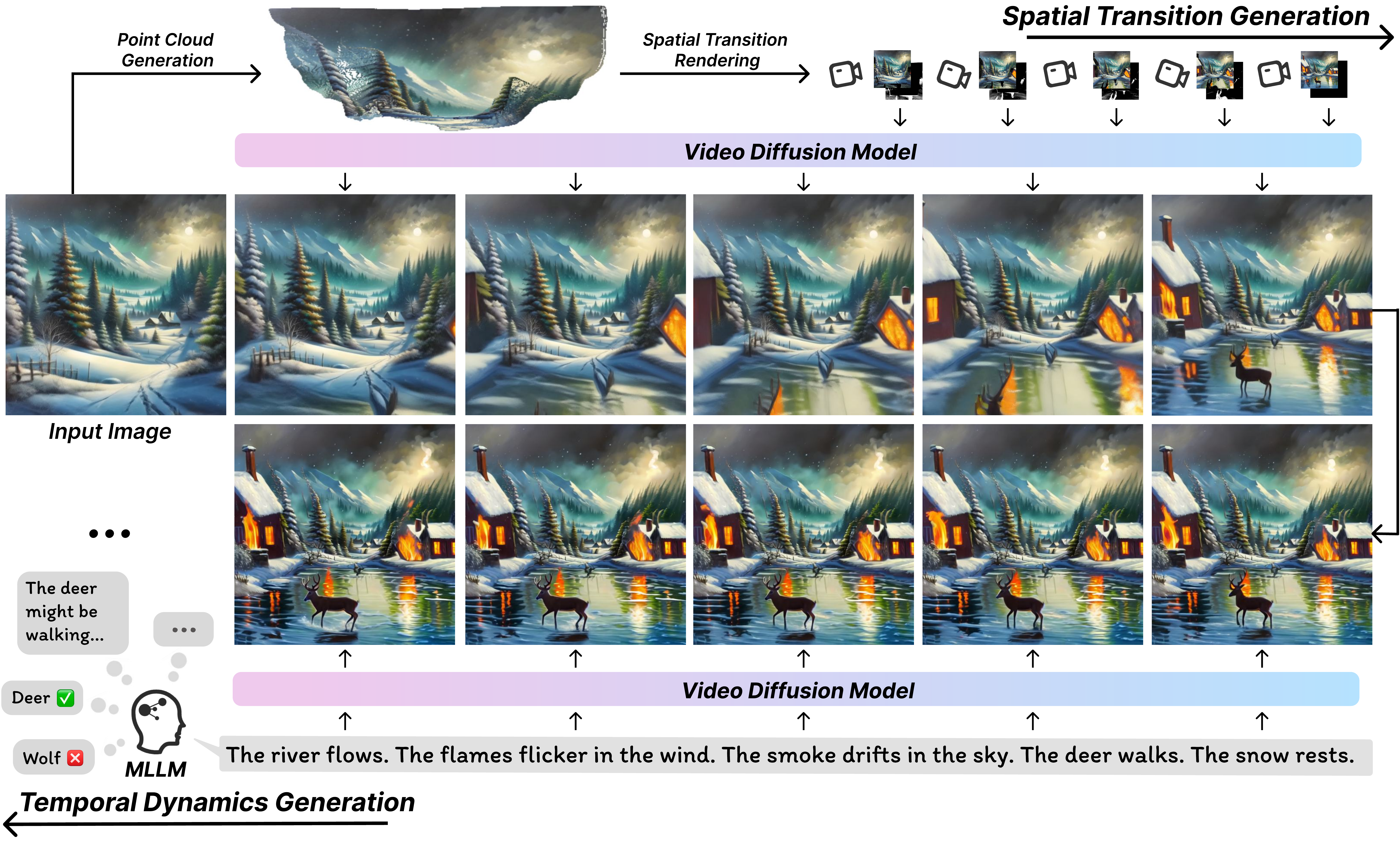} 
\caption{\textbf{Perpetual view generation.} Given a single input image, our method can perpetually generate a series of diverse yet coherent video clips depicting spatial transitions of the 3D scene and object dynamics in the scene. We highly encourage readers to check our project page for vivid video examples.}
\vspace{-0.2in}
\label{figs:intro}
\end{figure*}

\IEEEPARstart{I}{magine} you are looking at a single photo of a scene and suddenly find yourself transported into the world within the photo. As you step inside, the scene unfolds around you, transforming from a flat, still image into a vibrant, four-dimensional world. You move around, discovering new perspectives that were not visible in the original frame. When you pause to observe, you will see the scene come alive with subtle movements, such as the sway of grass in the wind or the gentle ripple of water in a nearby stream. You walk and pause, savoring each moment of this immersive journey. As humans, we have the extraordinary ability to imagine this journey in our minds. However, it is not trivial to mimic this for modern AI and computer vision models. 

In the early stage, some approaches \cite{Liu2020InfiniteNP, Li2022InfiniteNatureZeroLP} attempt to emulate the aforementioned imaginary capability by framing it as the problem of \emph{perpetual view generation}: synthesizing a long-term video corresponding to an arbitrary camera trajectory solely from a single input image. These methods often rely on large-scale image/video datasets for training and tend to be confined to a specific type of scene (e.g., landscapes \cite{Chai2023PersistentNA} or bedrooms \cite{ren2022look}).
Recently, the success of diffusion models \cite{Ho2020DenoisingDP, SohlDickstein2015DeepUL, yuan2024diffmat, zhong2025generative} has significantly advanced open-world image content creation \cite{Nichol2021GLIDETP, Ramesh2022HierarchicalTI, rombach2021highresolution}. Inspired by this, modern studies \cite{Fridman2023SceneScapeTC, yu2023wonderjourney} turn the focus to utilizing pre-trained 2D diffusion models to accomplish perpetual view generation in a zero-shot manner. These methods typically lift the input image to a 3D scene, represented by mesh \cite{Fridman2023SceneScapeTC} or point cloud \cite{yu2023wonderjourney}. Then the 2D diffusion model is used to fill in the previously unseen regions as the camera travels through the 3D scene. Despite exhibiting promising zero-shot perpetual view generation for various scenes, these 2D diffusion based approaches still suffer from unnatural results such as visual artifacts and geometric distortion, particularly along scene boundaries. The underlying rationale is that the pre-trained 2D diffusion model is trained solely on individual 2D images, thereby lacking 3D awareness and resulting in inconsistent 3D scene generation. In addition, all these works only synthesize videos depicting camera movements in static 3D scenes, failing to capture and represent dynamic object movements. This limitation significantly hinders the creation of an immersive experience that integrates both camera movements and object dynamics for users.  

In this paper, we propose to alleviate the above issues from the viewpoint of exploiting video diffusion models to trigger a new perpetual dynamic scene view generation problem. Unlike 2D diffusion models that lack 3D awareness and temporal dynamics, video diffusion models are trained with massive and various videos, including those featuring complex camera movements and object dynamics (e.g., drone footage and time-lapse photography). This enables video diffusion models to capture inherent cues from the dynamic world, and thus generate videos with great spatio-temporal consistency. Motivated by this, we present DreamJourney, a novel two-stage approach that fully excavates prior knowledge from the pre-trained video diffusion model to facilitate the perpetual dynamic scene view generation with high visual quality and motion consistency. Specifically, in stage \uppercase\expandafter{\romannumeral 1}, we lift the input image to a 3D point cloud by estimating its depth via a pre-trained monocular depth estimation model \cite{Ranftl2022} and render a sequence of partial images corresponding to a specific camera trajectory from the 3D point cloud. Then a pre-trained video diffusion model is utilized as generative prior to generate a high-quality video that adheres to the 3D scene and camera trajectory, as the diffusion sampling process is guided by the rendered sequential images. Meanwhile, we introduce two simple yet effective strategies (i.e., early stopping and view padding) to boost the video diffusion process with enhanced visual quality. Stage \uppercase\expandafter{\romannumeral 2} aims to add object dynamics to the scene. Specifically, we leverage a Multimodal Large Language Model (MLLM) to identify possible dynamic entities in the current view and generate a text prompt describing reasonable object movements of each entity. The resultant text prompt is then fed into the video diffusion model along with the current view to generate a following video with dynamic object movements. By repeating both stages recurrently, our DreamJourney allows perpetual view generation that contains both camera movements and object dynamics (see Figure \ref{figs:intro}).  

The main contribution of this work is the proposal of a two-stage video diffusion based paradigm for perpetual view generation with object dynamics, which is seldom investigated in the literature. This also leads to the elegant views of how to fully unleash the inherent world simulator power of pre-trained video diffusion models and how to properly arrange object dynamics with MLLM to animate scenes in this challenging task. Extensive experiments demonstrate that our DreamJourney outperforms state-of-the-art methods both quantitatively and qualitatively.

\section{Related Work}

\textbf{Perpetual View Generation} can be traced back to the pioneering work of Infinite Images \cite{Kaneva2010InfiniteIC}, which creates the effect of traversing an endless scene by retrieving, stitching, and rendering images from a database according to a predefined camera path. Inspired by this, Infinite Nature \cite{Liu2020InfiniteNP} and its follow-up works \cite{Li2022InfiniteNatureZeroLP, Chai2023PersistentNA} leverage GANs trained on domain-specific datasets to generate perpetual views on specific domains such as nature landscapes. Since the recent emergence of generative diffusion models trained on large-scale open-domain image datasets, subsequent works further expand the scope and applicability of this field. SceneScape \cite{Fridman2023SceneScapeTC} proposes a text-driven perpetual view generation approach with the aid of a cave-like lengthy mesh representation. WonderJourney \cite{yu2023wonderjourney} instead leverages LLMs and point cloud representation to generate perpetual views with more diversity and visual complexity than previous works. However, the perpetual views generated by WonderJourney lack object dynamics and contain noticeable artifacts during spatial transitions. Our work introduces video diffusion models into the perpetual view generation task for the first time to our best knowledge, infusing the process with dynamic elements and elevating the visual quality of spatial transitions. 

\textbf{Video Diffusion Models.}
Diffusion models \cite{Ho2020DenoisingDP, SohlDickstein2015DeepUL} have shown remarkable capabilities in image generation \cite{Nichol2021GLIDETP, Ramesh2022HierarchicalTI,rombach2021highresolution, T2I_Adapter} and 3D generation \cite{yang20233dstyle, chen2023control3d, yang2024dreammesh, chen2024vp3d, yang2024hi3d}. To replicate this success in the video domain \cite{chen2019mocycle, ho2022video, chen2019animating}, the first video diffusion model \cite{ho2022video} proposes to model video sequences \cite{wang2020story, wang2022coherent} using a 3D U-Net structure that is factorized over space and time \cite{xie2022global, song2022contextual, song2021spatial}. Then rapid progress has been made to enhance the quality and improve the training efficiency of video diffusion models \cite{ he2022lvdm, Blattmann2023StableVD, blattmann2023videoldm, danier2023ldmvfi}. Several works have also focused on image or video quality assessment \cite{huang2023vbench, UIQI, UnderwaterIQA, BlindQA}.
Recently, Sora \cite{OpenAI2023VideoGeneration} and its open-source counterparts \cite{opensora, ma2024latte, xu2024easyanimatehighperformancelongvideo} have significantly improved the generated video quality and length by adopting more scalable transformer architectures \cite{pan2025stream, yao2024hiri} in place of 3D U-Nets. In this work, we propose a modular framework to harness the rapidly growing generative power of video diffusion models for the new task of perpetual dynamic scene view generation.

\begin{figure*}[t]
\centering
\includegraphics[width=1.0\textwidth]{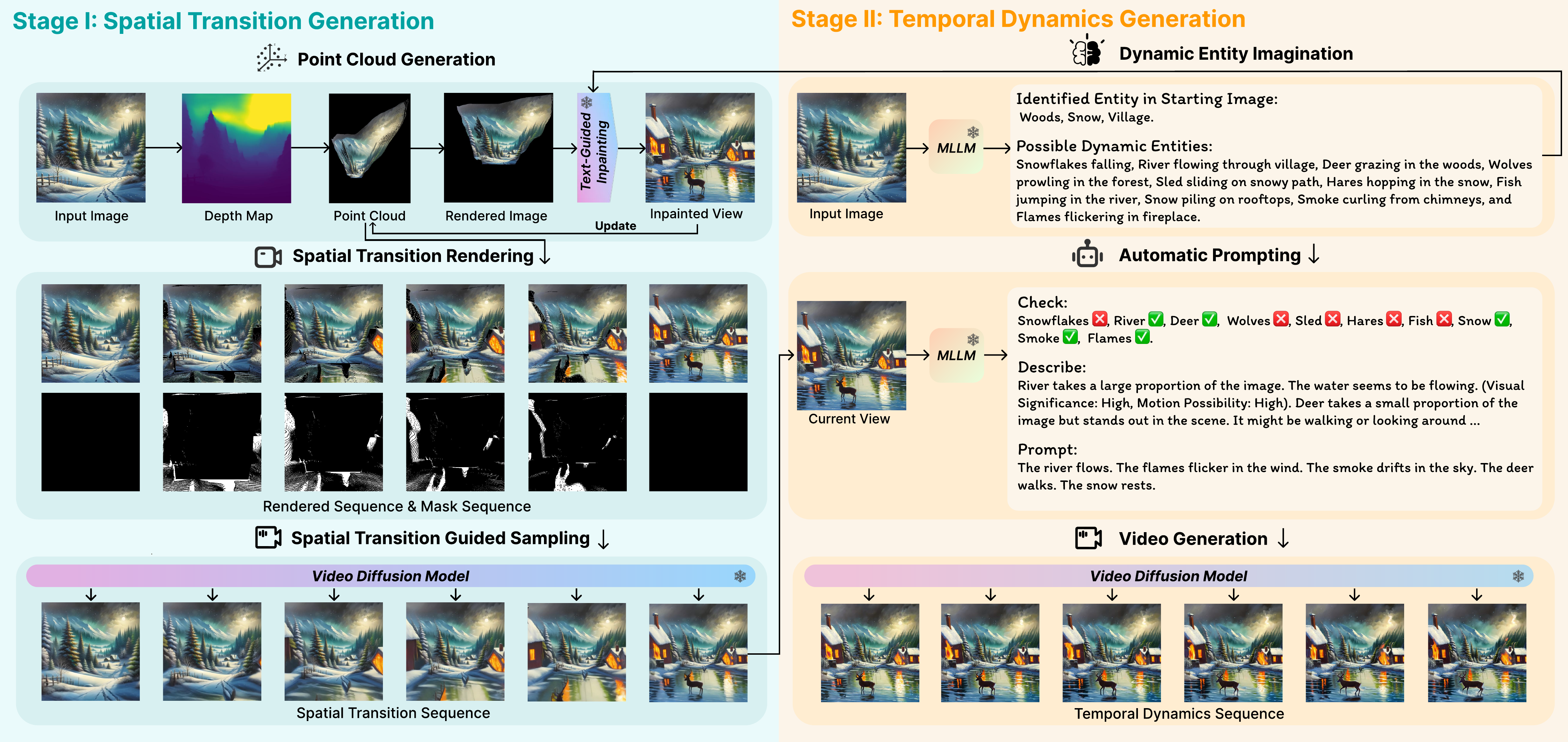} 
\caption{An overview of our DreamJourney framework, which novelly leverages
the world simulation capacity of video diffusion models to trigger a new perpetual dynamic scene view generation task. DreamJourney consists of two main stages: spatial transition generation and temporal dynamics generation. In stage \uppercase\expandafter{\romannumeral 1}, DreamJourney lifts the input image to a 3D point cloud and renders a sequence of partial images corresponding to a specific camera trajectory. Then these rendered images are used to guide the video sampling process of a pre-trained video diffusion model in a diffusion posterior sampling perspective, resulting in a high-quality 3D scene spatial transition video. In stage \uppercase\expandafter{\romannumeral 2}, DreamJourney utilizes MLLM as an automatic prompting agent to write a text prompt describing scene object dynamics in a chain-of-thought style. This text prompt is then fed into the video diffusion model to animate the scene. By alternately repeating stage \uppercase\expandafter{\romannumeral 1} and \uppercase\expandafter{\romannumeral 2}, DreamJourney allows perpetual dynamic scene view generation that contains both camera movements and object dynamics.}
\vspace{-0.2in}
\label{fig_framework}
\end{figure*}

\section{Preliminary---Diffusion Models}
We first briefly review the diffusion models \cite{Ho2020DenoisingDP}, which are latent-variable generative models that learn to gradually transform a sample from a tractable noise distribution towards a target data distribution. Diffusion models consist of a \emph{diffusion process} and a \emph{reverse process}. In the diffusion process, diffusion models gradually add Gaussian noise to the clean data $\mathbf{x}_0  \sim p_{data}(\mathbf{x}_0)$ according to a predefined variance schedule $\beta_1,...,\beta_T $:
\begin{equation}
\label{eq:diffusion_process}
    q(\mathbf{x}_{1:T}|\mathbf{x}_0) := \prod_{t=1}^{T}q(\mathbf{x}_t|\mathbf{x}_{t-1}),
\end{equation}
\begin{equation}
\label{eq:diffuse}
    q(\mathbf{x}_t | \mathbf{x}_{t-1}) := \mathcal{N}(\mathbf{x}_t; \sqrt{1 - \beta_t} \mathbf{x}_{t-1}, \beta_t\mathbf{I}).
\end{equation}
Finally, the data point $\mathbf{x}_T$ becomes indistinguishable from pure Gaussian noise. It also admits sampling $\mathbf{x}_t$ at an arbitrary timestep $t$ in closed form:
\begin{equation}
    q(\mathbf{x}_t | \mathbf{x}_{0}) := \mathcal{N}(\mathbf{x}_t; \sqrt{\bar{\alpha}_t} \mathbf{x}_{0}, (1-\bar{\alpha}_t)\mathbf{I}),
\end{equation}
where $\alpha_t = 1 - \beta_t, \bar{\alpha}_t = \prod_{s=1}^t \alpha_s$.
In the reverse process, diffusion models are trained to reverse the diffusion process, i.e., recovering $\mathbf{x}_0$ from $\mathbf{x}_T$, which is defined as:
\begin{equation}
\label{eq:diffusion_process}
    p_{\theta}(\mathbf{x}_{0:T}) := \prod_{t=1}^{T}p_{\theta}(\mathbf{x}_{t-1}|\mathbf{x}_{t}),
\end{equation}
\begin{equation}
    p_{\theta}(\mathbf{x}_{t-1} | \mathbf{x}_{t}) := \mathcal{N}(\mathbf{x}_{t-1}; \boldsymbol{\mu}_{\theta}(\mathbf{x}_t,t), \boldsymbol{\Sigma}_{\theta}(\mathbf{x}_t,t)).
\end{equation}
In practice, $\boldsymbol{\mu}_{\theta}(\mathbf{x}_t,t)$ is commonly parameterized as:
\begin{equation}
  \begin{aligned}
    \boldsymbol{\mu}_{\theta}(\mathbf{x}_t,t) = \frac{1}{\sqrt{\alpha_t}} 
    \Big( \mathbf{x}_t - \frac{\beta_t}{\sqrt{1 - \bar{\alpha}_t}} 
    \boldsymbol{\epsilon}_\theta(\mathbf{x}_t,t) \Big),
  \end{aligned}
\end{equation}
where $\boldsymbol{\epsilon}_\theta$ is a a neural network with parameter $\theta$. The $\boldsymbol{\Sigma}_{\theta}(\mathbf{x}_t,t)$ is simply set as $\beta^2_{t}\mathbf{I}$. Then the training objective of diffusion models can be simplified as follows:
\begin{equation}
  \mathcal{L}_\mathrm{simple}(\theta) := \|\boldsymbol{\epsilon}_\theta(\mathbf{x}_t,t) - \boldsymbol{\epsilon} \|^2_2,
\label{equation:diffloss}
\end{equation}
where $\boldsymbol{\epsilon}_\theta$ is trained to predict the added noise $\boldsymbol{\epsilon}$ from $\mathbf{x}_t$. After training, diffusion models can randomly sample a noise vector $\mathbf{x}_T$ and gradually denoise it until it reaches a high-quality output data $\mathbf{x}_0$. 


\section{DreamJourney}

In this section, we present our proposed DreamJourney, which novelly leverages video diffusion models to tackle the new challenging task of perpetual dynamic scene view generation. Our launching point is to exploit the intrinsic world simulation capability of the video diffusion models to generate continuous views with plausible camera movements and object dynamics. Figure \ref{fig_framework} illustrates an overview of the DreamJourney framework, which consists of two main stages: spatial transition generation and temporal dynamics generation. We will elaborate on the details of these two stages in the following subsections. 

\subsection{Stage \uppercase\expandafter{\romannumeral 1}: Spatial Transition Generation}
In this stage, given a single RGB image $I_0$ and its camera pose $C_0$, we aim to synthesize a video clip $\{I_i\}^{N}_{i=0}$ corresponding to a camera trajectory $\{C_i\}^{N}_{i=0}$ that depicts a flythrough of the scene captured by the initial view $I_0$, where $C_i \in \mathcal{R}^{3\times4}$ is the camera pose of the $i^{th}$ frame $I_i$ in the synthesized video clip.

\textbf{Point Cloud Generation.} To fulfill this goal, we first lift the input image to a 3D point cloud that represents the underlying 3D scene. Thus we can precisely model the camera movements in the 3D space. We first leverage an off-the-shelf monocular depth estimation model \cite{Ranftl2022} to predict the depth map $D_0$ of the input image $I_0$ and use this depth map to unproject $I_0$ into a point cloud $\mathcal{P}$ with a perspective camera model:
\begin{equation}
    \mathcal{P} = \phi(I_0,D_0,C_0,K),
\end{equation}
where $\phi$ is the image to point cloud unprojection function and $K$ is the camera intrinsic matrix. Given this initial point cloud, we render a partial image $\hat{I}_N$ with a mask $M_N$ indicating unseen regions from camera pose $C_N$, which is the endpoint of the camera trajectory $\{C_i\}^{N}_{i=0}$ and is the farthest from the initial view. Then we use a pre-trained text-guided inpainting model \cite{rombach2021highresolution} to fill the unseen regions in $\hat{I}_N$ conditioned on a text prompt $Y_N$. We leverage an MLLM to generate the text prompt $Y_N$ describing reasonable objects that might appear in $\hat{I}_N$. To be clear, we ask MLLM to first identify existing entities in the input view $I_0$ and then propose $k$ possible entities in the nearby scene that harmonize with the input view and these existing entities. In addition, since we aim to generate perpetual views with both camera movements and object dynamics in this work, we encourage the MLLM to put entities with larger motion possibilities in front positions of the text prompt. Thus the rendered image $\hat{I}_N$ tends to be filled in moving objects, which is beneficial for the temporal dynamics generation in the next stage. We then unproject the inpainted image $I_N$ back into 3D and merge it with the initial point cloud $\mathcal{P}$ to obtain a relatively complete point cloud $P^{*}$, which is used for the following spatial transition video generation. 

\textbf{Spatial Transition Rendering.} After obtaining the point cloud $\mathcal{P}^*$, we can render a sequence of partial images $\{\hat{I}_i\}^{N-1}_{i=1}$ and visible masks $\{M_i\}^{N-1}_{i=1}$ along the camera trajectory $\{C_i\}^{N-1}_{i=1}$. To produce the spatial transition video, one simplest way is to inpaint these partial images $\{\hat{I}_i\}^{N-1}_{i=1}$ by an image inpainting model, according to the visible masks $\{M_i\}^{N-1}_{i=1}$ that distinguishes the unseen regions from existing ones. However, this inevitably leads to misalignment across different inpainted images, making the resultant video temporally inconsistent. To alleviate this issue, WonderJourney chooses to gradually inpaint the partial image sequence in the camera trajectory by iteratively performing 2D image inpainting, depth estimation, and point-cloud merging \cite{yu2023wonderjourney}. Nevertheless, this method still suffers from visual artifacts and temporal inconsistency, as the 2D image inpainting model is not aware of the whole 3D transition sequence and the depth prediction error is prone to accumulate in the frequent depth estimation process.

\textbf{Spatial Transition Guided Sampling.}
Unlike previous methods \cite{Fridman2023SceneScapeTC, yu2023wonderjourney}, we pave a new way to synthesize the spatial transition video by utilizing a pre-trained video diffusion model. Specifically, instead of directly inpainting the partial image sequence, we novelly treat them as the scene spatial transition prior to guide the sampling process of the video diffusion model, which can produce a video that not only precisely adheres to the 3D scene and camera movements but also has great visual quality and temporal consistency. We employ the open-sourced video diffusion model EasyAnimate\cite{xu2024easyanimatehighperformancelongvideo}, which is capable of sampling video conditioned on a starting frame $I_s$, and a text prompt $Y$. Formally, the video sampling process starts from a randomly sampled noise video $\mathbf{x}_T \sim \mathcal{N}(\mathbf{0}, \mathbf{I})$, which is then iteratively denoised by $T$ time steps to produce the final clean video $\mathbf{x}_0$. In each time step $t$, the denoiser $\boldsymbol{\epsilon}_\theta$ in the pre-trained video diffusion model predicts the added noise of $\mathbf{x}_t$ and then denoise $\mathbf{x}_t$ to $\mathbf{x}_{t-1}$:
\begin{equation}
  \begin{aligned}
    \mathbf{x}_{t-1} = \frac{1}{\sqrt{\alpha_t}} 
    \Big( \mathbf{x}_t - \frac{\beta_t}{\sqrt{1 - \bar{\alpha}_t}} 
    \boldsymbol{\epsilon}_\theta(\mathbf{x}_t,t,
    I_s, Y) \Big) + \sigma_t \mathbf{z},
  \end{aligned}
\end{equation}
where $\mathbf{z} \sim \mathcal{N}(\mathbf{0}, \mathbf{I})$ and $\sigma^{2}_t=\beta_t$. We set input view $I_0$ as $I_s$ and a fixed description \textit{``a camera flyover, cruising steadily across the scene"} as the text prompt $Y$, Now we introduce how we inject the scene appearance and camera movement cues behind the rendered partial image sequence into this video sampling process. First, we formulate a prior video $\mathbf{x}^p$ that consists of a sequence of images $\{I_0, \hat{I}_1, ...,\hat{I}_{N-1}, I_N\}$, where the first frame is the input view, the last frame is the inpainted view and the intermediate frames are the rendered partial images. Meanwhile, we have a corresponding mask video $\mathbf{m}^p$ that consists of $\{M_0, M_1, ...,M_{N-1}, M_N\}$, which indicates visible regions of each frame in $\mathbf{x}^p$. The prior video $\mathbf{x}^p$ indicates the spatial transition of the 3D scene, which motivates us to utilize it to guide the video sampling process of the pre-trained video diffusion model. Specifically, at each time step $t$, we can estimate the final clean sample $\hat{\mathbf{x}}_0(\mathbf{x}_t)$ from $\mathbf{x}_t$ by:
\begin{equation}
  \begin{aligned}
    \hat{\mathbf{x}}_0(\mathbf{x}_t) = \big( \mathbf{x}_t - \sqrt{1 - \bar{\alpha}_t}
    \boldsymbol{\epsilon}_\theta(\mathbf{x}_t,t, I_s, Y) \big).
  \end{aligned}
\end{equation}
Then we guide the video diffusion sampling by adding a gradient descent step on $\mathbf{x}_{t-1}$ according to prior video $\mathbf{x}^p$:
\begin{equation}
\label{eq:dps}
  \begin{aligned}
    \hat{\mathbf{x}}_{t-1} = \mathbf{x}_{t-1} -s_t \nabla_{\mathbf{x}_t} \| \hat{\mathbf{x}}_0(\mathbf{x}_t) \odot \mathbf{m}^p - \mathbf{x}^p \odot \mathbf{m}^p \|^2_2,
  \end{aligned}
\end{equation}
where $s_{t}$ denotes the guidance weight. Intuitively, the second term  $\nabla_{\mathbf{x}_t} \| \hat{\mathbf{x}}_0(\mathbf{x}_t) \odot \mathbf{m}^p - \mathbf{x}^p \odot \mathbf{m}^p \|^2_2$ in Eq. (\ref{eq:dps}) provides a update direction of $\mathbf{x}_t$ to push the estimated final clean video $\hat{\mathbf{x}}_0(\mathbf{x}_t)$ closer to the prior video $\mathbf{x}^p$ in visible regions. Different from directly blending the visible region of $\mathbf{x}^p$ into the $\mathbf{x}_{t}$ at each time step, Eq. (\ref{eq:dps}) achieves the prior guided sampling harmoniously in a diffusion posterior sampling perspective. To further improve the semantic consistency between the visible and invisible regions of the sampled video, we also apply a renoising step that diffuses the output $\hat{\mathbf{x}}_{t-1}$ back to $\mathbf{x}_t$ by using Eq. (\ref{eq:diffuse}) in the video sampling process \cite{lugmayr2022repaintinpaintingusingdenoising}.

Furthermore, we propose two simple but effective strategies to improve the quality of the generated video. The first one is early stopping. Since the diffusion denoising process tends to establish large-scale structures in the early time steps and refine small-scale details in the later time steps, we propose to stop injecting the video prior to the video sampling process after time step $t_{c}$, which allows the pre-trained video diffusion model to fully unleash its generative power. The second one is view padding. Note that the input view $I_0$ and inpainted view $I_N$ have more awareness of the 3D scene $\mathcal{P}^*$ than the rendered partial images. We replicate them in the prior video during the video sampling process to strengthen the scene information in video generation. Then we remove duplicate ones in the final generated video.

\subsection{Stage \uppercase\expandafter{\romannumeral 2}: Temporal Dynamics Generation}
In stage \uppercase\expandafter{\romannumeral 1}, we synthesize a video $\{I_i\}^{N}_{i=0}$ flying through a scene adhering to a specific camera trajectory. By iteratively repeating stage \uppercase\expandafter{\romannumeral 1}, we can achieve perpetual view generation. However, the resultant perpetual views only contain spatial transitions, lacking object dynamics, which severely limits the immersive experience. In light of this, we propose to generate temporal object dynamics in this stage. Specifically, we take the last frame $I_N$ of the generated video in stage \uppercase\expandafter{\romannumeral 1} as the starting view in this stage. One simplest way is to directly use $I_N$ as the starting frame condition and use a default text guidance (e.g. \textit{``The video is of high quality, high dynamic, and the view is very clear..."}) to prompt the video generation model. However, the generated video has a large possibility of containing large camera motion and static objects. That is because when the video generation models are not informed with text guidance about what is the specific dynamics in the video, they tend to generate a ``safer" result - more camera motion and less object dynamics.

To tackle this, we propose to use MLLM with vision capability to play the role of an automatic prompting agent for video diffusion models. Specifically, we propose a chain-of-thought \cite{wei2023chainofthoughtpromptingelicitsreasoning} style three-step prompting strategy. First, we prompt MLLM to crosscheck the objects in possible dynamic entities description $Y_N$ and the entities in the inpainted image $I_N$ to identify the entities that occur both in $I_N$ and $Y_N$. Second, we require MLLM to observe the image $I_N$ to describe the visual significance and possible motion of each object identified in the first step. Third, we prompt MLLM to write a final prompt $Y_D$ for the dynamic scene based on the description in the second step. 
Based on $I_N$ as the starting image and the atomically generated prompt $Y_D$ as the guiding text, we can leverage the video diffusion model to generate a video sequence with desired object dynamics.
Herein we employ the same video generation model \cite{xu2024easyanimatehighperformancelongvideo} as used in stage \uppercase\expandafter{\romannumeral 1}. By alternately repeating stage \uppercase\expandafter{\romannumeral 1} and stage \uppercase\expandafter{\romannumeral 2}, we can generate a potentially infinite-length video that depicts a perpetual view sequence containing both spatial transition and object dynamics.

\section{Experiments}
\begin{figure*}[!t]
\centering
\includegraphics[width=1.0\textwidth]{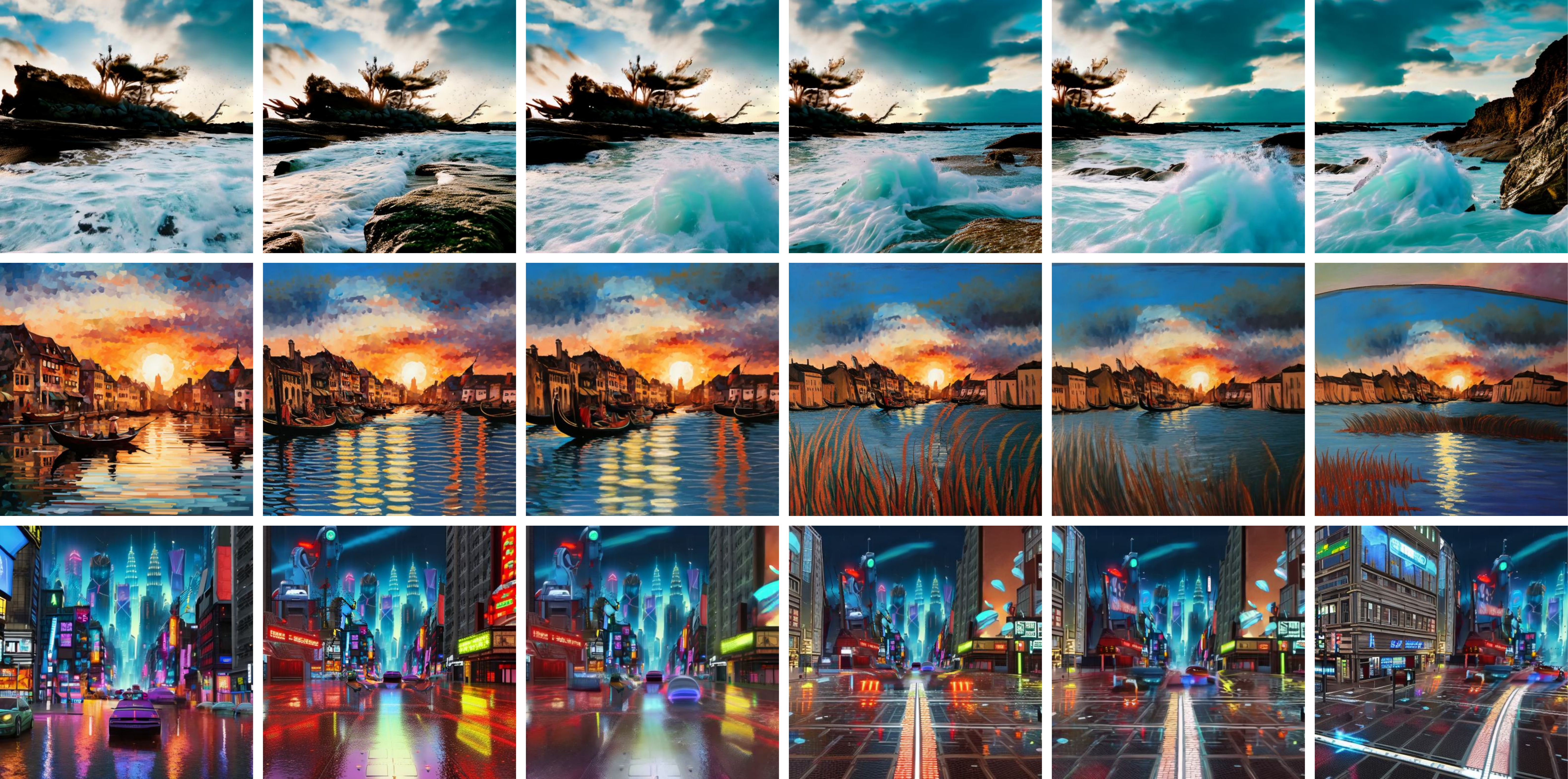} 
\caption{Qualitative results for immersive journeys generated by our DreamJourney, showing that DreamJourney can generate long-range views with plausible scene transitions and object dynamics. \textbf{We recommend visiting our project page for a demonstration of full trajectories aligned with WonderJourney.}}
\vspace{-0.2in}
\label{fig_quality_demo}
\end{figure*}

\begin{figure}[!t]
\centering
\includegraphics[width=1.0\columnwidth]{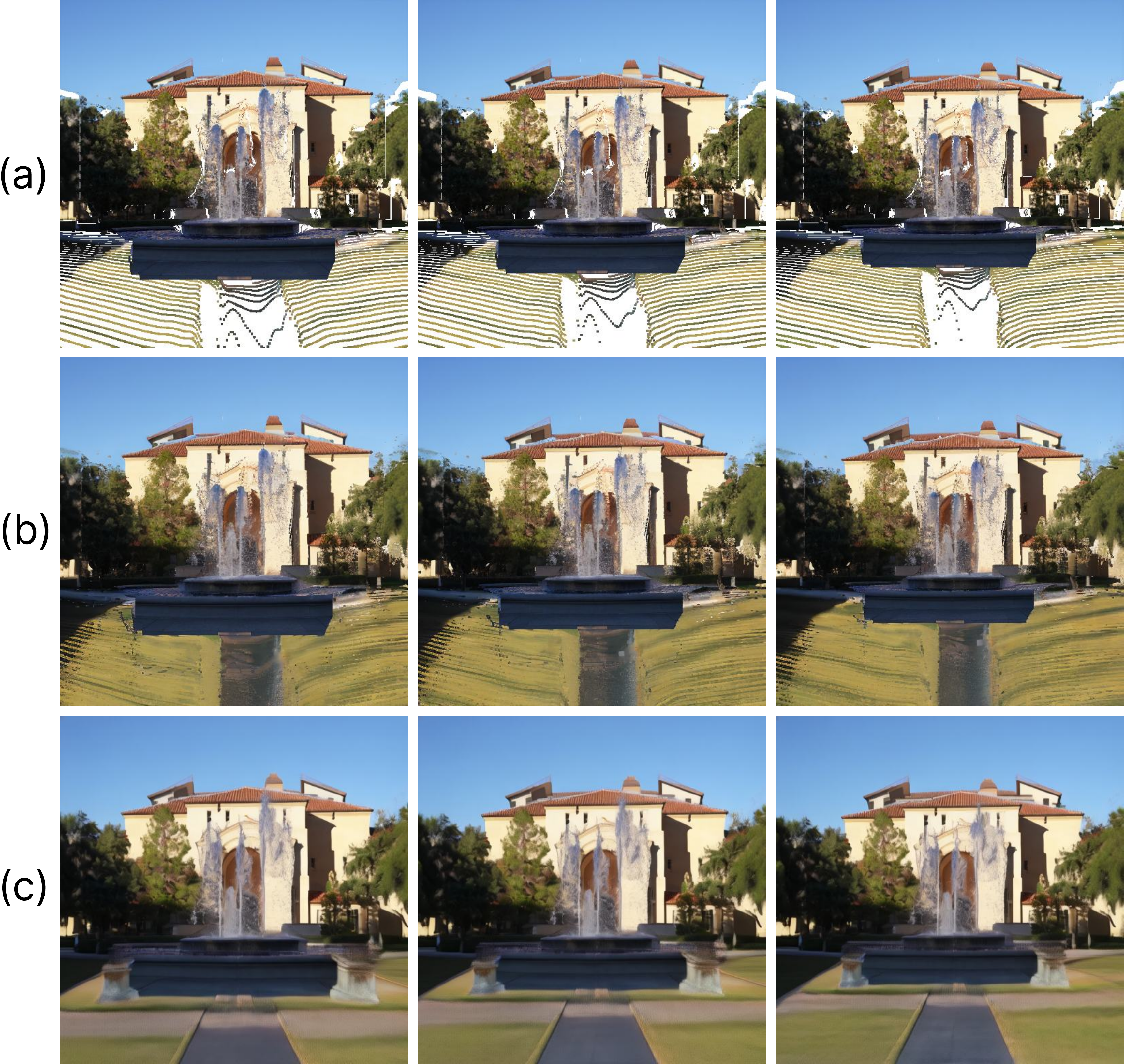} 
\caption{Qualitative analysis of the effect of spatial transition guided sampling. (a) input partial images, (b) outputs by iterative 2D inpainting and point cloud merging, and (c) outputs by our proposed spatial transition guided sampling.}
\vspace{-0.2in}
\label{fig_spatialAblation}
\end{figure}

\begin{figure*}[!t]
\centering
\includegraphics[width=0.95\textwidth]{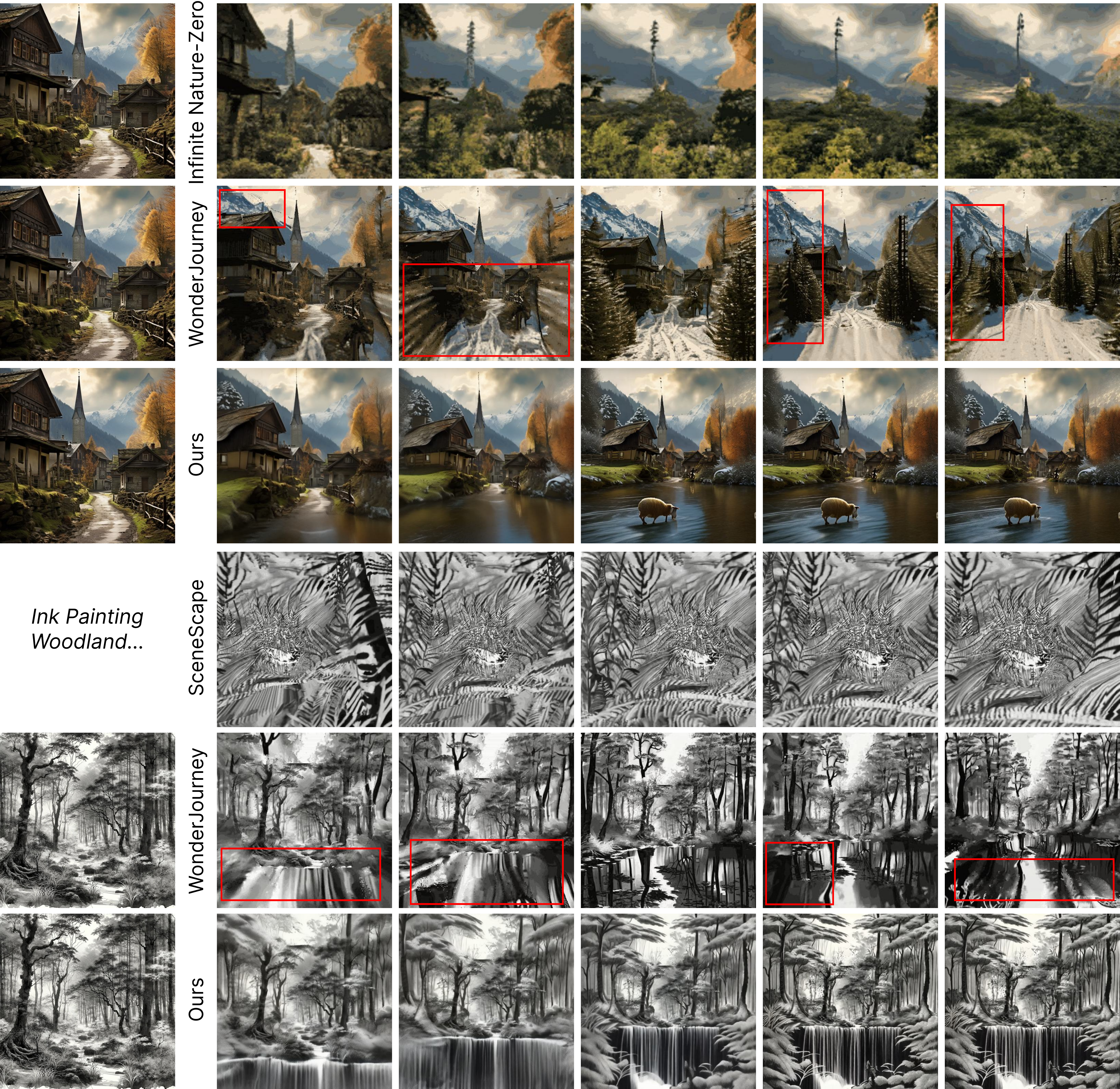} 
\caption{Qualitative comparisons with InfiniteNature-Zero, SceneScape, and WonderJourney.}
\vspace{-0.2in}
\label{fig_quality_compare}
\end{figure*}

\subsection{Dataset and Baselines}
We evaluate our method on the dataset collected by WonderJourney \cite{yu2023wonderjourney}, which consists of real-world or AI-generated images within a diverse range of themes (e.g. real-world photos, ink paintings). We compare our method against three state-of-the-art perpetual view generation methods:  Infinite Nature-Zero \cite{Li2022InfiniteNatureZeroLP}, SceneScape \cite{Fridman2023SceneScapeTC}, and WonderJourney.

\subsection{Implementation Details}
In Stage \uppercase\expandafter{\romannumeral 1}, we use MiDaS v3.1 \cite{Ranftl2022} for monocular depth estimation. The camera paths for spatial transition generation, utilized in our experiments, adhere to the methodology outlined in Wonderjourney \cite{yu2023wonderjourney}. Specifically, for linear motion, we implemented a backward camera movement of 0.0005 units, with a random sine
perturbation to the height of the intermediate cameras. For rotational transitions, we applied a rotation of 0.45 radians combined with a translational movement of 0.0001 units. For spatial transition guided sampling, the early stop rate is set as 20\% (i.e. skip injecting the rendered video prior during the last 20\% time steps), and the round of re-noising conducted during the sampling process is set as 15. The number of padded views in the view padding strategy is set as 4. We use EasyAnimate v3 \cite{xu2024easyanimatehighperformancelongvideo} and set the output frames as 48 for both spatial transition generation and temporal dynamics generation. We use GPT-4 \cite{openai2024gpt4technicalreport} as the multi-model large language model during experiments. All experiments are conducted on a symmetric multiprocessing node with AMD EPYC 7763 64-Core Processors and NVIDIA A100-SXM4-80GB GPU.

\subsection{Experimental Results and Analysis}

\textbf{Qualitative demonstration.} 
Herein we showcase three qualitative examples of the generated journeys in Figure \ref{fig_quality_demo}. These results demonstrate that our DreamJourney is capable of generating an immersive journey that contains smooth scene transitions and vivid object dynamics. Take the first row in Figure \ref{fig_quality_demo} for instance, from left to right, the camera first zooms out, and the rocks appear in the scene. Then the camera stops moving, we see the waves crashing against the rocks. Next, the camera moves to the right, continuing the journey. We can observe that these generated views are harmoniously coherent and consistent with the properties of the physical world. This indicates that our DreamJourney can potentially create fantastic virtual experiences for users.

\textbf{Qualitative comparisons.}
We depict the qualitative comparisons in Figure \ref{fig_quality_compare}. As illustrated in the figure, Infinite Nature-Zero is prone to generate morphing transitions with less spatial and temporal consistency. SceneScape tends to generate cave-like scenes due to its cave-like lengthy mesh representation, which is unsuitable for expansive outdoor scenes (e.g. woodland). Compared with these two methods, WonderJourney achieves higher visual quality. However, it still suffers from generating artifacts during spatial transitions, especially along scene boundaries. In contrast, our DreamJourney manages to produce high-quality results that are harmoniously coherent with the input view. 


\textbf{User study.}
We conduct a user study to quantitatively compare our DreamJourney against the baseline methods. We follow WonderJourney's evaluation procedure and metrics. We show participants two videos side-by-side with randomized positions in each test case. One of them is the video generated by DreamJourney and another one is generated by a baseline method. We then ask participants to choose the better one for each comparison by considering four different aspects: (1) the diversity of scenes within a single journey, (2) visual quality, (3) scene complexity, and (4) overall interestingness. According to all participants’ feedback, we measure the user preference score of one method as the percentage of its generated results that are preferred. Table \ref{tbl:human} shows the results of the user study. In general, our DreamJourney significantly outperforms the baseline methods with higher user preference rates in all measuring aspects.

\begin{table}[!t]
    \centering
    \caption{Human preference user study. We compare our DreamJourney with baseline methods on diversity, visual quality, scene complexity, and overall interestingness.}
    \label{tbl:human}
    \renewcommand\arraystretch{1.0}
    \resizebox{0.47 \textwidth}{!}{
    \begin{tabular}{l|cccc}
    \hline
      Comparison   & Div. & Qual. & Compl. & Overall  \\
         \midrule
      DreamJourney \emph{vs.} InfiniteNature-Zero  & 92.5\% & 94.5\% & 91.8\% & 96.3\%  \\
      DreamJourney \emph{vs.} SceneScape & 92.3\% & 93.3\% & 91.0\% & 94.3\% \\
      DreamJourney \emph{vs.} WonderJourney & 80.8\% & 90.3\% & 81.2\% & 93.5\% \\

      \hline
    \end{tabular}}
\end{table}

\begin{table}[!t]
    \centering
    \tiny    
    \caption{Human preference user study. We compare our DreamJourney with ablated methods on visual quality.}
    \label{tbl:ablation_study}
    \renewcommand\arraystretch{1.0}
    \resizebox{0.4 \textwidth}{!}{
    \begin{tabular}{l|cccc}
    \hline
      Comparison  &  Qual. \\
         \midrule
      DreamJourney \emph{vs.} w/o Early Stopping &  89.3\% &  \\
      DreamJourney \emph{vs.} w/o View Padding &  92.3\% & \\

      \hline
    \end{tabular}}
\vspace{-0.2in}
\end{table}

\begin{figure}[t]
\centering
\includegraphics[width=1.0\columnwidth]{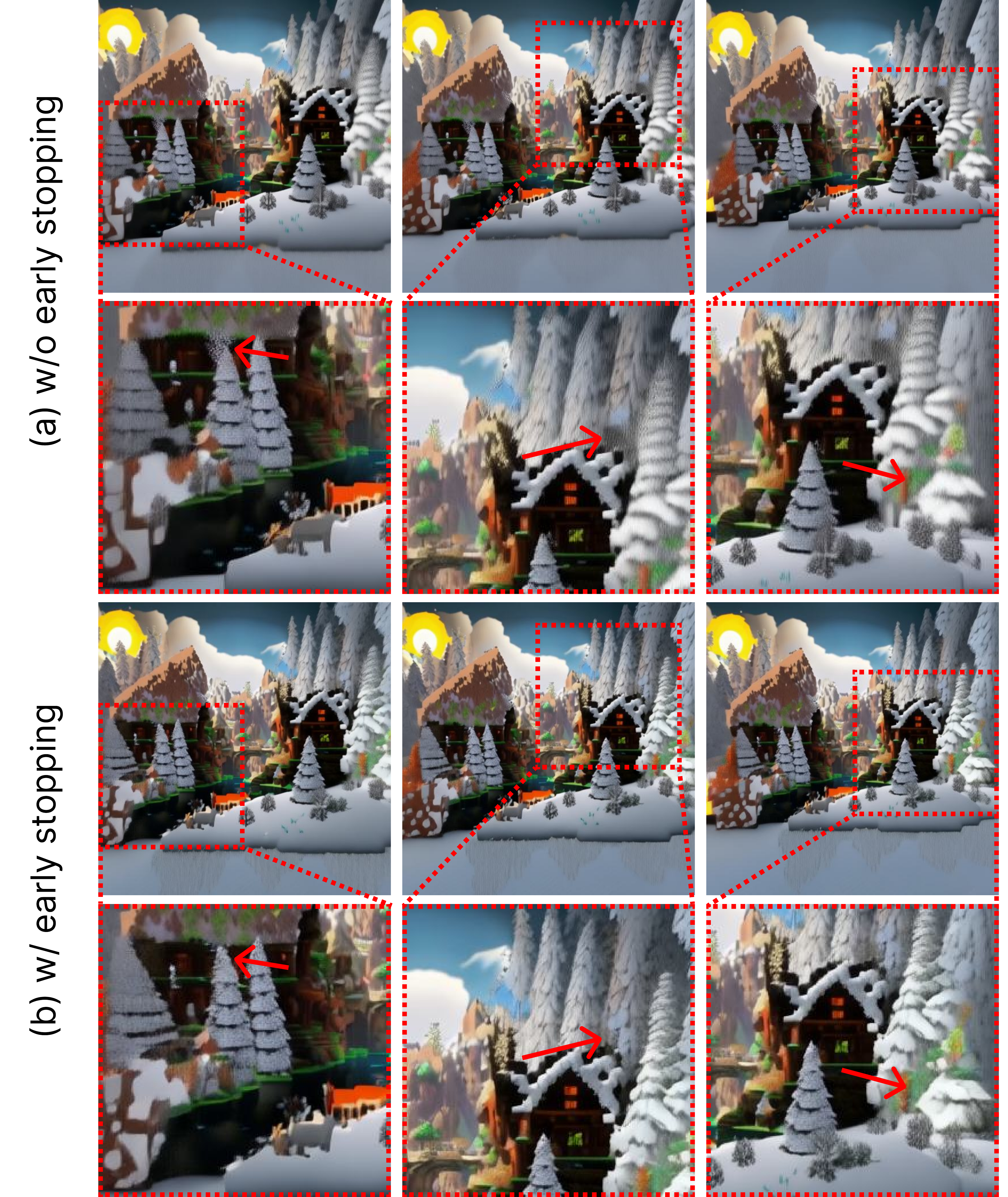} 
\caption{Ablation study on the proposed early stopping strategy. Without early stopping, the generated frames tend to contain artifacts or blurs around object boundaries (see areas pointed by red arrows). With early stopping, these artifacts can be corrected with the generative prior of video diffusion models.}
\label{spatialAblation_earlyStop}
\end{figure}

\begin{figure}[t]
\centering
\includegraphics[width=1.0\columnwidth]{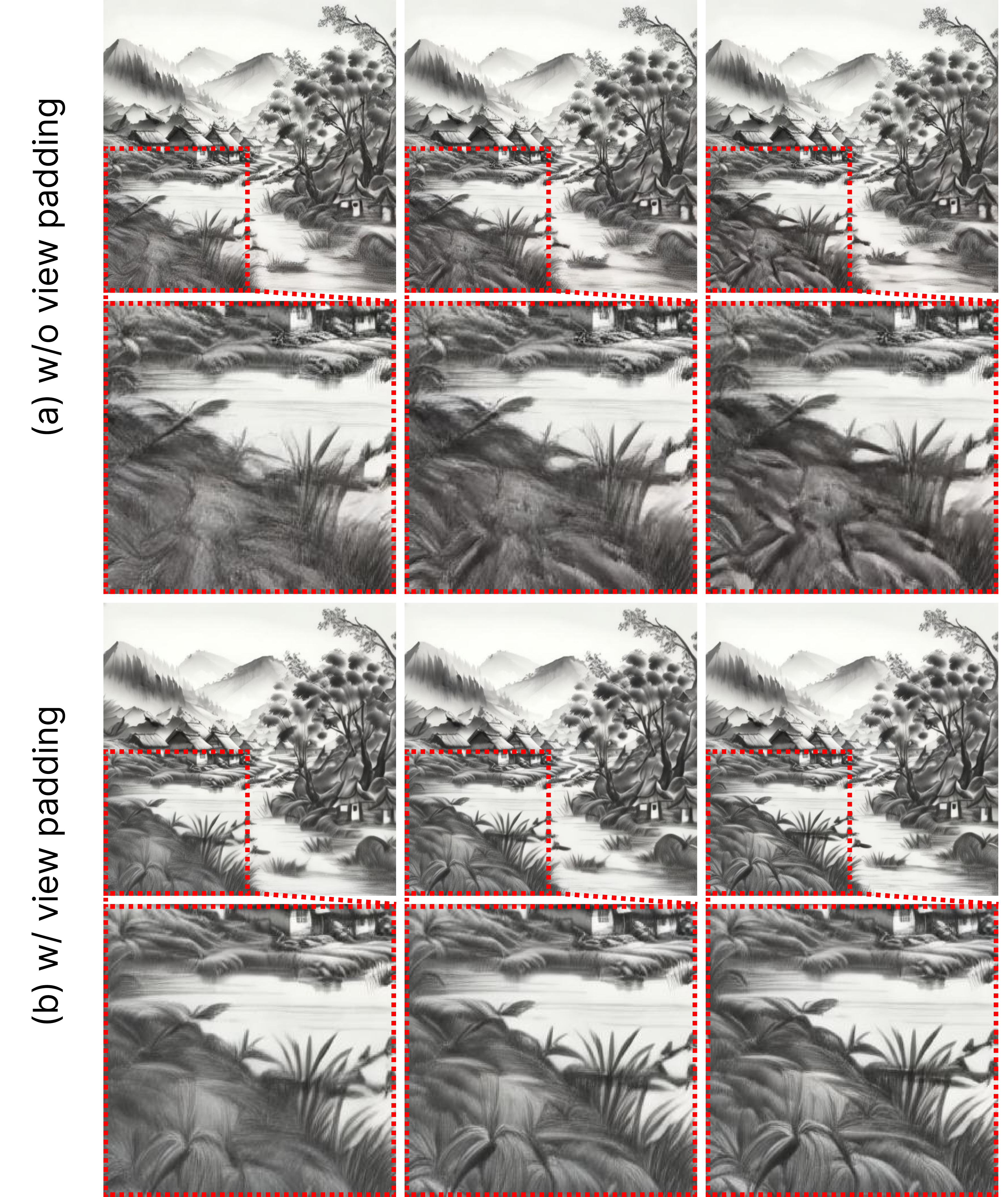} 
\caption{Ablation study on the proposed view padding strategy. It can be seen that the view padding strategy helps better preserve areas with rich detail, such as the grass in the lower-left corner of the views.}
\label{spatialAblation_viewPadding}
\end{figure}

\subsection{Ablation Studies}
In this part, we investigate the effectiveness of the proposed spatial transition guided sampling in stage \uppercase\expandafter{\romannumeral 1} and the proposed two strategies (i.e., early stopping and view padding).

\textbf{Effect of spatial transition guided sampling.}
Given the rendered partial image sequence (Figure \ref{fig_spatialAblation} (a)), WonderJourney chooses to iteratively perform 2D image inpainting, depth estimation, and point-cloud merging to inpaint the partial image sequence (Figure \ref{fig_spatialAblation} (b)). In this way, the inpainted regions are consistent across different views. However, the inpainted sequence contains obvious artifacts. For instance, as shown in Figure \ref{fig_spatialAblation} (b), the grassland has unnatural strips that originate from the input sequence. Moreover, the generated pavement at the center is out of harmony with its surroundings. In contrast, our DreamJourney leverages pre-trained video diffusion models and uses the rendered partial image sequence as video prior to guide the video diffusion sampling process, resulting in more plausible results. As depicted in Figure \ref{fig_spatialAblation} (c), the proposed spatial transition guided sampling manages to produce a video sequence that precisely inherits the scene appearance and camera movements of input partial images. At the same time, it can correct the artifacts in the input sequence and obtain more harmonious results. These results validate the effectiveness of our proposed spatial transition guided sampling. 

\textbf{Effect of early stopping.} 
While the prior video obtained by point cloud rendering indicates the spatial transition of the 3D scene, it contains unnatural artifacts caused by inaccurate point location estimation, especially around object boundaries. If we let the prior video guide the video generation of the whole diffusion denoising process, the generated video may inherit these artifacts or blurs in the corresponding area (e.g. the areas pointed by red arrows in Figure \ref{spatialAblation_earlyStop} (a)). Since the diffusion denoising process tends to establish large-scale structures in the early time steps and refine small-scale details in the later time steps, we can stop injecting the video prior during later time steps to fully unleash the generative power of video diffusion models to address those artifacts. As shown in Figure \ref{spatialAblation_earlyStop} (b), those artifacts can be corrected by stopping using the video prior guidance in the last twenty percent time steps, which validates the effectiveness of our proposed early stopping strategy. The quantitative results in Table \ref{tbl:ablation_study} further support this observation.

\textbf{Effect of view padding.} 
Since the starting and ending views for spatial transition generation provide more comprehensive contexts of the scene than the rendered partial images, we duplicate these views in the prior video to enhance the scene information during spatial transition-guided sampling. We then remove those duplicated views before finalizing the video. In this context, we compare the final video with and without the application of the view padding strategy. As shown in Figure \ref{spatialAblation_viewPadding}, using the view padding strategy in the video diffusion sampling process helps to better preserve areas with rich detail, such as the grass in the lower-left corner of the views. This validates the effectiveness of our proposed view padding strategy in the spatial transition guided sampling, as further supported by the quantitative results in Table \ref{tbl:ablation_study}.

\subsection{Additional Qualitative Results}
\begin{figure*}[tp]
\centering
\includegraphics[width=1.0\textwidth]{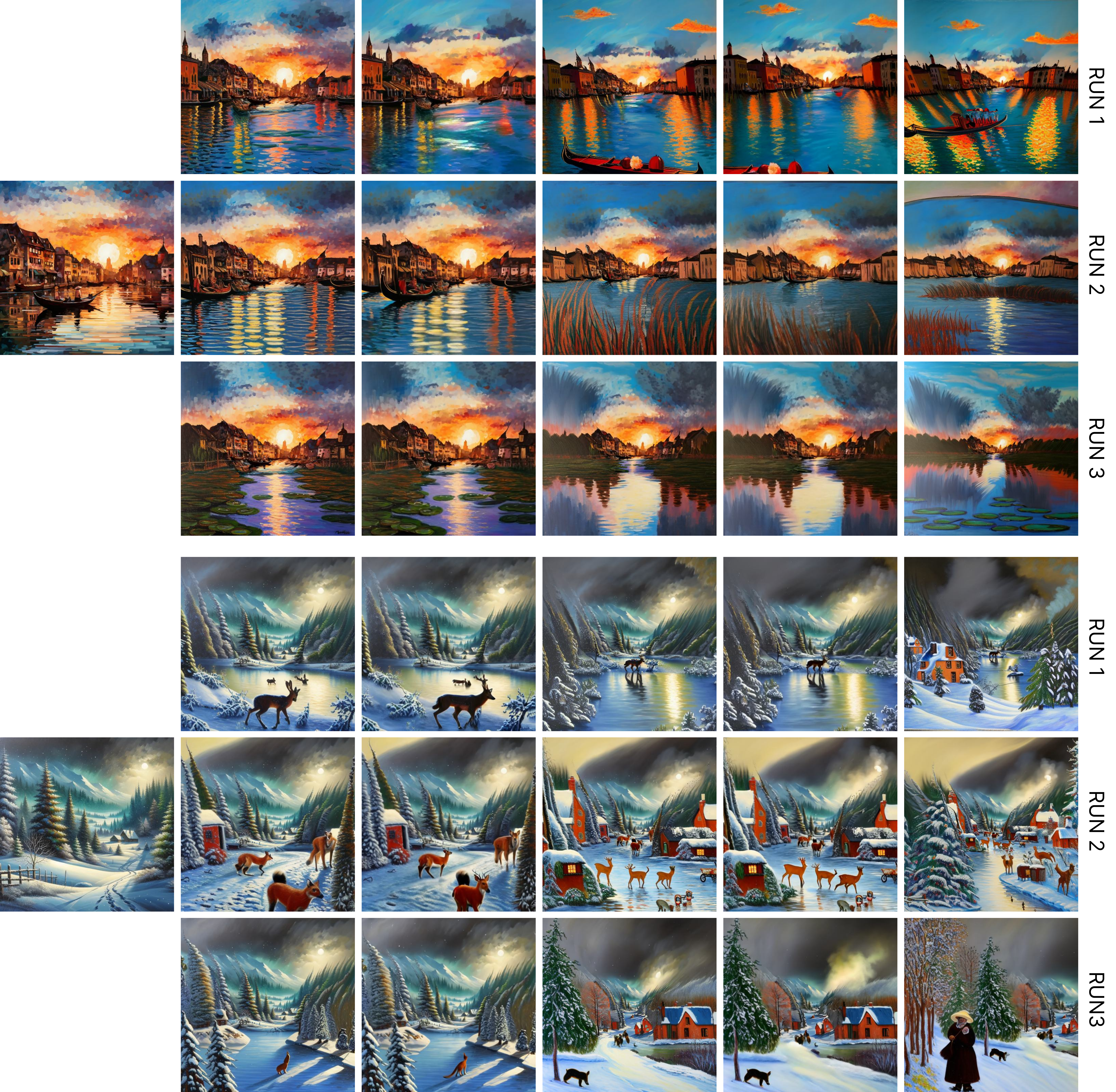} 
\caption{Qualitative results for diverse “dream journeys” generated by our DreamJourney from the same input image.}
\label{Appendix_Diversity}
\end{figure*}

\begin{figure*}[tp]
\centering
\includegraphics[width=1.0\textwidth]{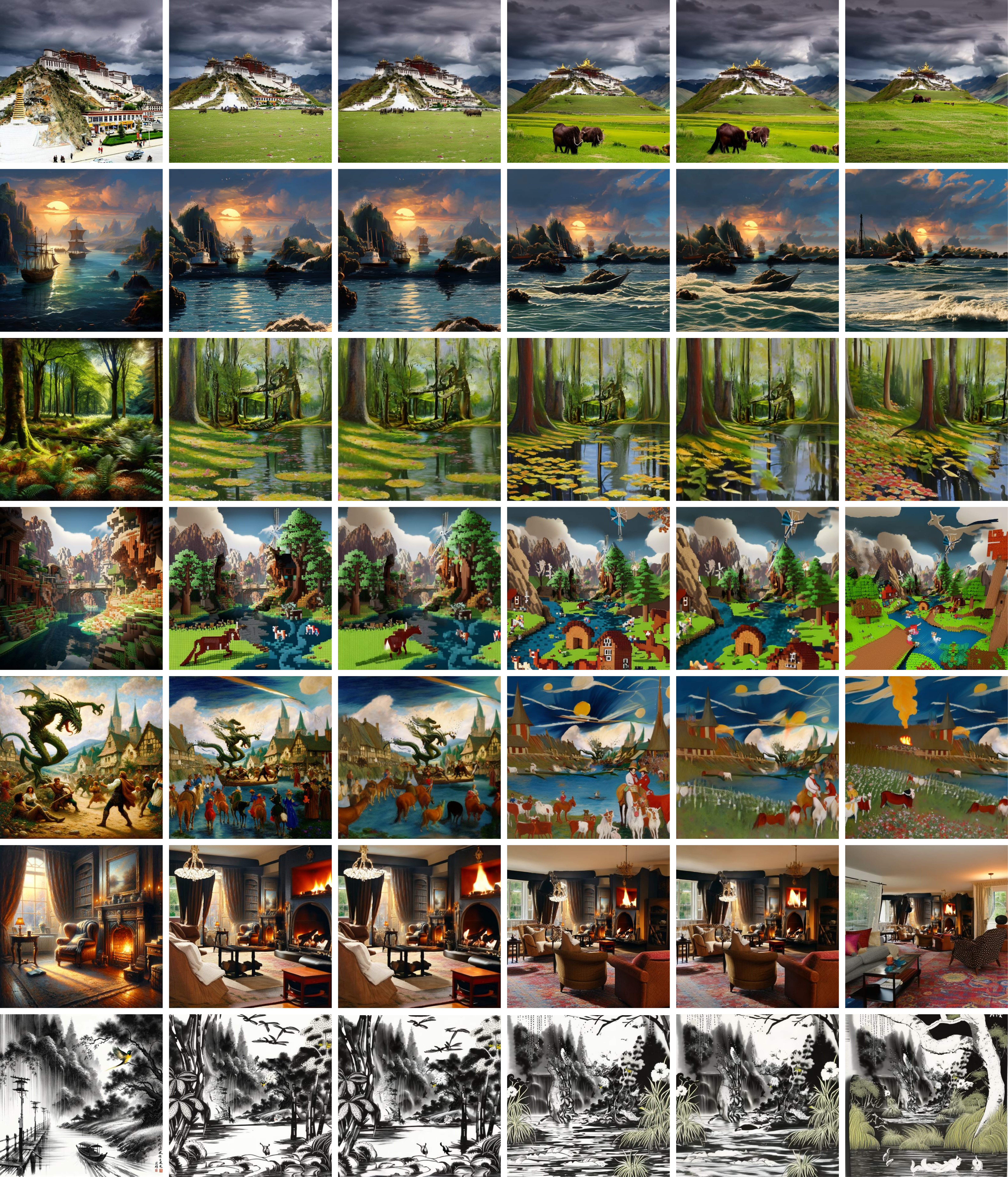} 
\caption{Additional qualitative results for various ``dream journeys'' generated by our DreamJourney. Despite the input images having diverse styles and entities, our DreamJourney can consistently generate immersive results.}
\label{figs:Appendix_QualityDemo}
\end{figure*}

We show additional qualitative examples of the generated journeys by our DreamJourney in Figure \ref{figs:Appendix_QualityDemo}. Moreover, our method can generate diverse journeys from the same input. We show these diverse examples in Figure \ref{Appendix_Diversity}. We highly encourage readers to check our project page \url{https://dream-journey.vercel.app/} for vivid video
examples.


\section{Conclusion}
We have presented DreamJourney, a two-stage framework that unleashes the inherent prior knowledge in the pre-trained video diffusion model to trigger a new perpetual dynamic scene view generation task. DreamJourney also involves the Multimodal Large Language Model (MLLM) as an automatic prompting agent to arrange object dynamics of scenes in a chain-of-thought style. The extensive experiments demonstrate that DreamJourney can generate fascinating long-term videos with vivid spatial transitions and object dynamics solely from a single input image, enabling users to craft their ``dream journeys'' with immersive experiences.

\section{Acknowledgments}
This work is supported by the National Natural Science Foundation of China (No. U22B2034, 62421003), Zhejiang Provincial Natural Science Foundation of China (No. LD24F020011), Beijing Municipal Science and Technology Project (No. Z241100001324002) and Beijing Nova Program (No. 20240484681). The authors would like to thank the anonymous reviewers for their valuable comments.


{\appendix[]
\section*{Prompt Design for the MLLM}
We use GPT-4 as the MLLM for both dynamic entity imagination and automatic prompting. For dynamic entity imagination, we first ask GPT-4 to identify current entities in the input image. Based on the current entity list, we then instruct GPT-4 as follows to obtain the dynamic entities description for later views:

\textit{``You are an intelligent scene generator. Imagine you are flying through a scene, based on the entities in the current scene, you need to imagine possible entities with common and significant visible motion in the scene. You need to generate a suitable scene name and description for the 10 main entities in the scene. The entities within the scenes are adapted to match and fit with the scenes, and you should put entities with larger visual significance and motion possibility first.''}

An example output for the scene name and generated dynamic entities description would be:

\textit{``Dragon Temple Garden: Dragon wings fluttering, sakura petals dropping, incense smoke swirling, lion statue's eyes blinking, flag rustling in the wind, koi fish swimming in the pond, lotus flower opening, owl hooting on the temple, beads on prayer wheels turning, and paper lanterns swaying gently.''}

For the automatic prompting stage, we instruct GPT4 to take a three-stage chain-of-thought style \cite{wei2023chainofthoughtpromptingelicitsreasoning} inference as follows to obtain the final dynamic description prompt for video generation:

\textit{
``\textbf{First stage (identify)}: You should check the objects in the list of possible dynamic objects one by one to verify if they exist in the given image. If there is no object in the list identified in the given image, then you need to identify some by yourself. }

\textit{\textbf{Second stage (describe)}: For each identified object, provide a concise description of the Visual Significance (i.e. the proportion in given image), Motion Possibility (i.e. possibility of containing strong motion in the next few seconds), and what motion it/they may have in the image. }

\textit{\textbf{Third stage (write prompt)}: Based on the descriptions from the previous stage, write the final dynamical description for the scene, first describe those objects with strong visual significance and motion possibility.''}

An example output for the chain-of-thought process and the final dynamical description would be:

\textit{
  ```Think Log': `First stage: 1. Dragon wings are identified in the given image. 2. Sakura petals are not identified in the given image. 3. Incense smoke is identified in the given image. 4. Lion statue's eyes are not identified in the given image. 5. Flag is not identified in the given image. 6. Koi fish is not identified in the given image. 7. Lotus flower is not identified in the given image. 8. Owl is not identified in the given image. 9. Beads on prayer wheels are not identified in the given image. 10. Paper lanterns are identified in the given image. Second Stage: 1. Dragon wings are spread and take a reasonable portion of the image. The wings seem ready to flap any moment. (Visual Significance: High, Motion Possibility: High). 2. Incense smoke is located in the background and takes a small portion of the image. The smoke seems to be wafting upwards. (Visual Significance: Low, Motion Possibility: High). 3. Paper lanterns are scattered throughout the scene and take a noticeable portion of the image. They seem to be gently swinging or floating. (Visual Significance: Medium, Motion Possibility: Medium). Third stage: The visual summary for dynamical description is 'The dragon's wings flap. The smoke wafts. The lanterns swing.'',}
  
  \textit{`Dynamical Description': `The dragon's wings flap. The smoke wafts. The lanterns swing.'"}

  \section*{Additional Analysis}
\begin{figure*}[tp]
\centering
\includegraphics[width=1.0\textwidth]{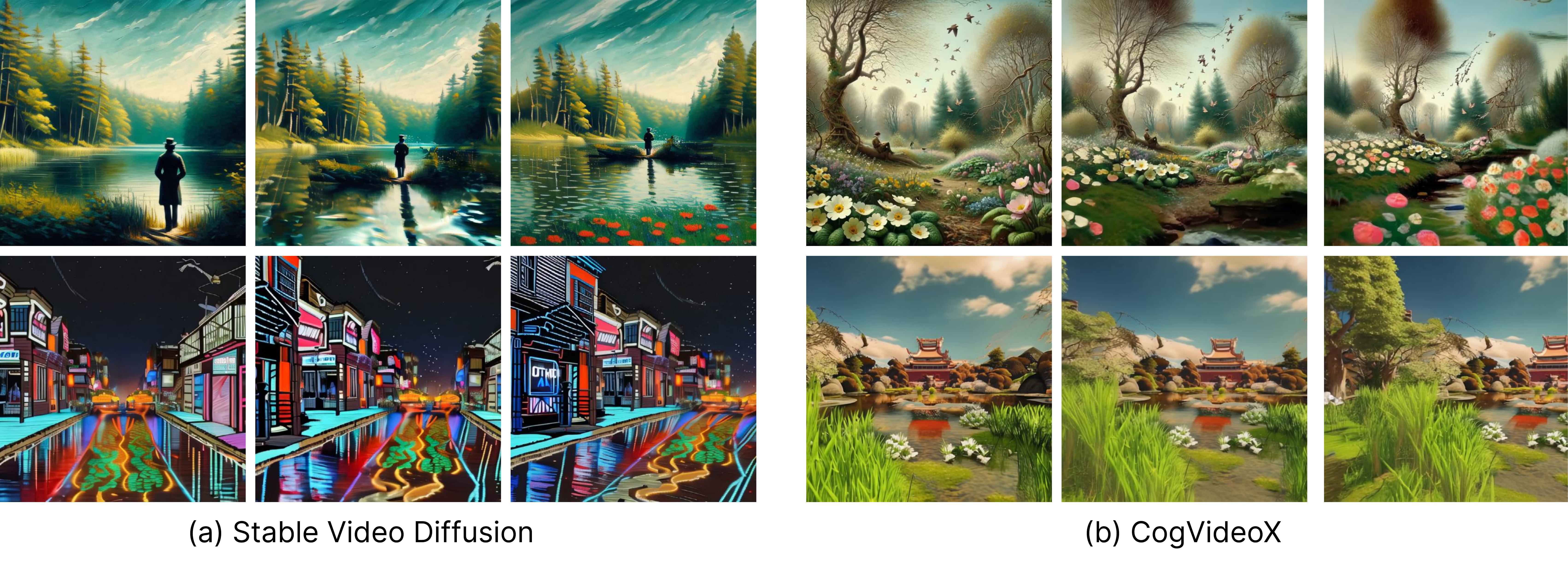} 
\caption{Framework generality demonstrated across different video diffusion foundation models. Stable Video Diffusion \cite{Blattmann2023StableVD}, CogVideoX \cite{yang2024cogvideox}, and EasyAnimate (default model) \cite{xu2024easyanimatehighperformancelongvideo} all successfully function within our framework for spatial transition generation, highlighting the framework's adaptability to various video generation models.}
\label{figs:different_videoModel}
\end{figure*}

\begin{figure*}[tp]
\centering
\includegraphics[width=1.0\textwidth]{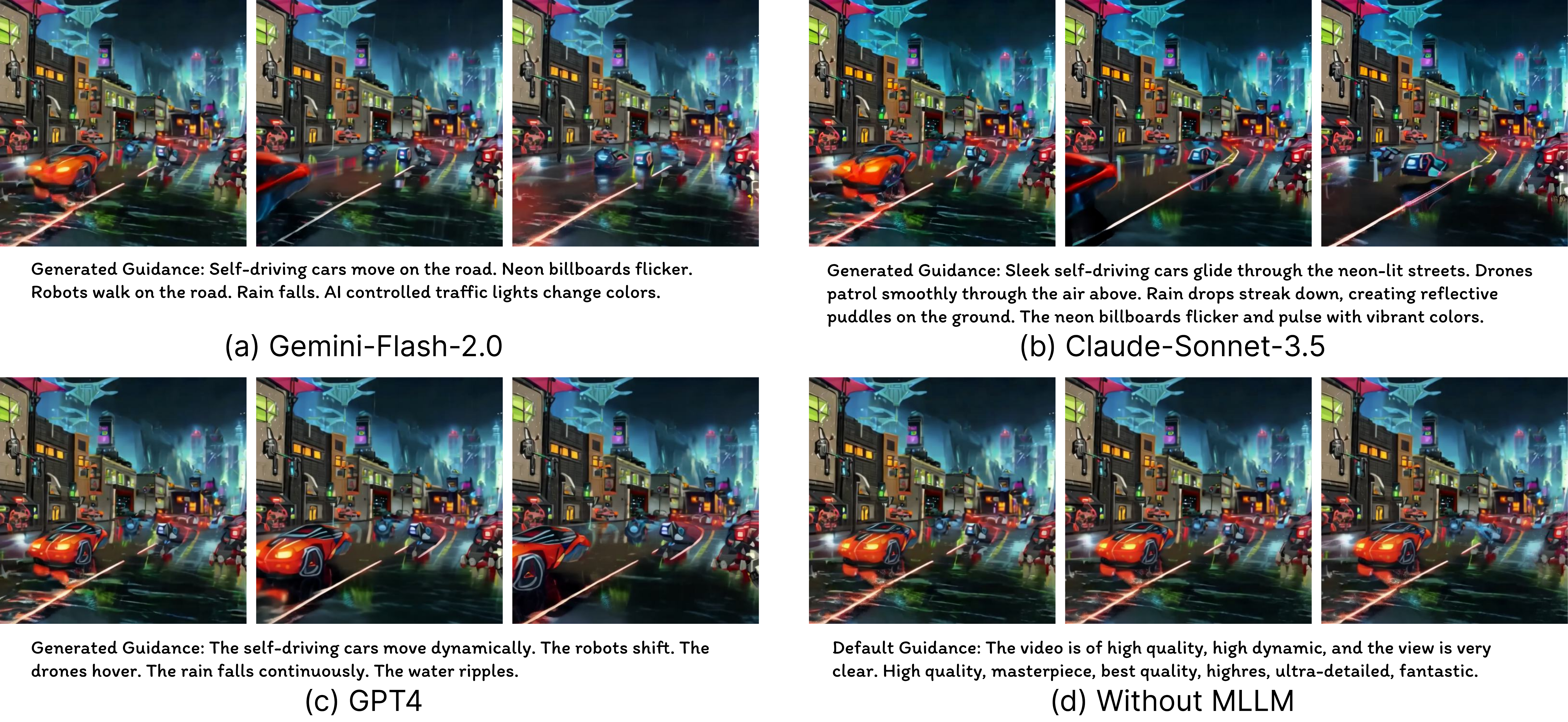} 
\caption{Comparison of video generations guided by different MLLMs. While each MLLM produces prompts with distinct stylistic characteristics, all successfully provide dynamic guidance for the scene, as evidenced by the movement of the red car across all MLLM-guided generations compared to its stationary position in the default guidance without MLLM intervention.}
\label{figs:different_MLLM}
\end{figure*}

\begin{figure}[t]
\centering
\includegraphics[width=1.0\columnwidth]{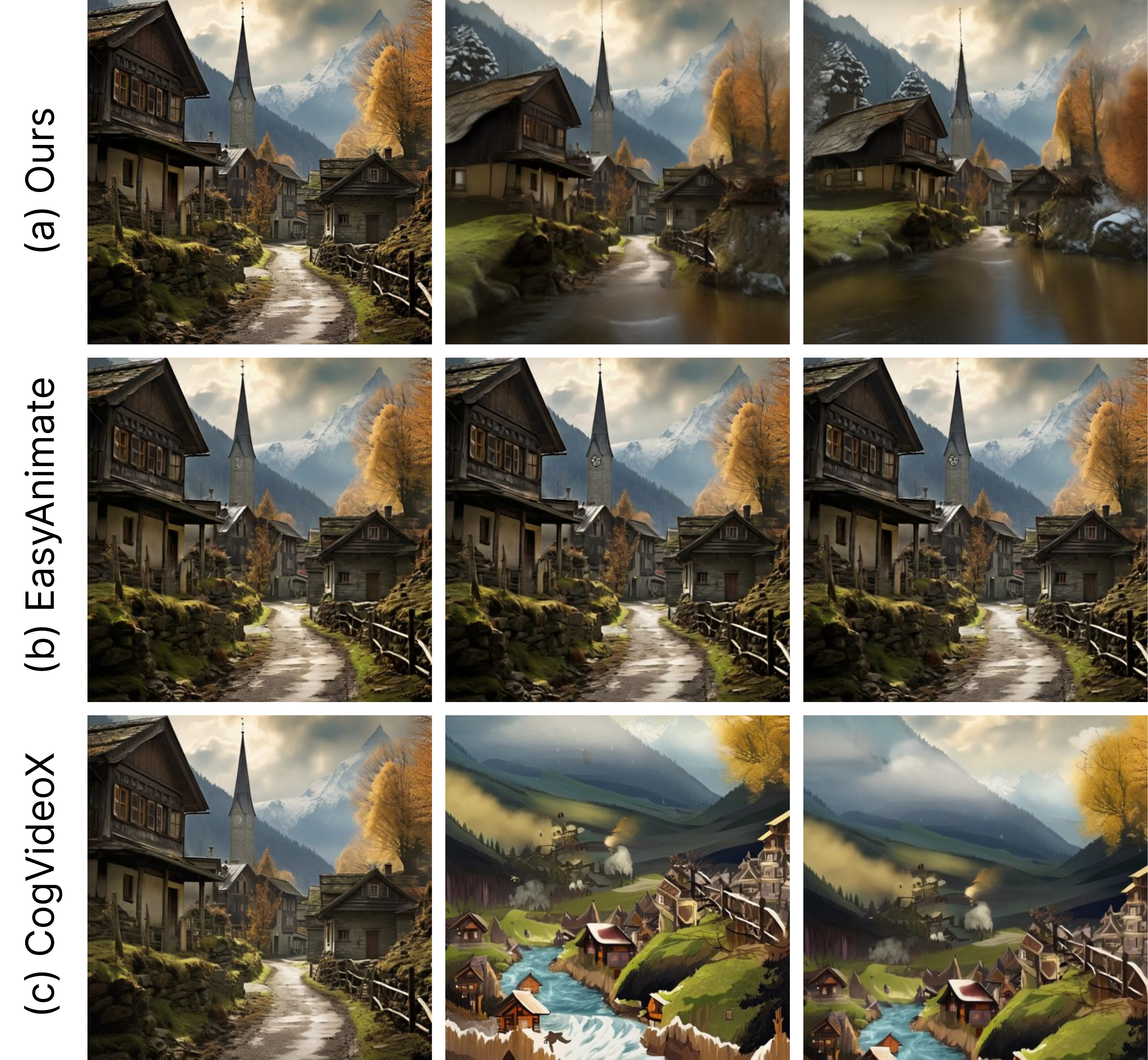} 
\caption{Comparison of our method with general video generation models. (a) Our method produces stable and substantial camera movements. (b) EasyAnimate \cite{xu2024easyanimatehighperformancelongvideo} shows minimal movement despite the "The camera zooms out steadily." requirement in prompt. (c) CogVideoX \cite{yang2024cogvideox} generates a montage cut instead of the required smooth camera transition.}

\label{fig: comparison_with_general_video_model}
\end{figure}

\begin{figure}[t]
\centering
\includegraphics[width=1.0\columnwidth]{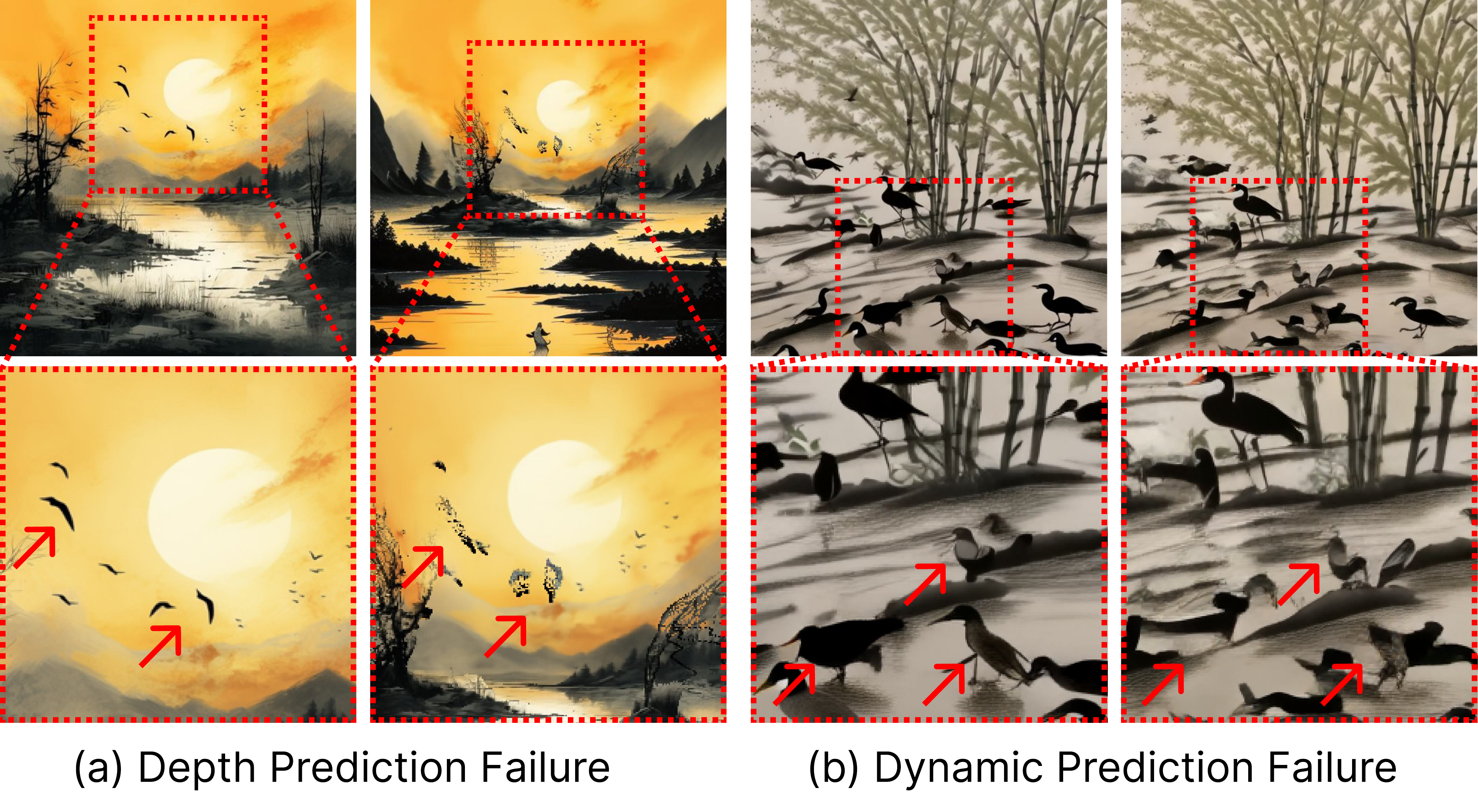} 
\caption{Typical failure cases. (a) Due to inaccurate depth prediction with current pre-trained depth estimation models, small distant birds are incorrectly reprojected as fragmented points in the 3D point cloud, resulting in visual artifacts. (b) Due to limitations in current pre-trained video generation models, complex scenes with multiple interacting birds lead to unrealistic movement patterns and object deformations.}

\label{fig: failure_case}
\end{figure}

\begin{figure*}[tp]
\centering
\includegraphics[width=1.0\textwidth]
{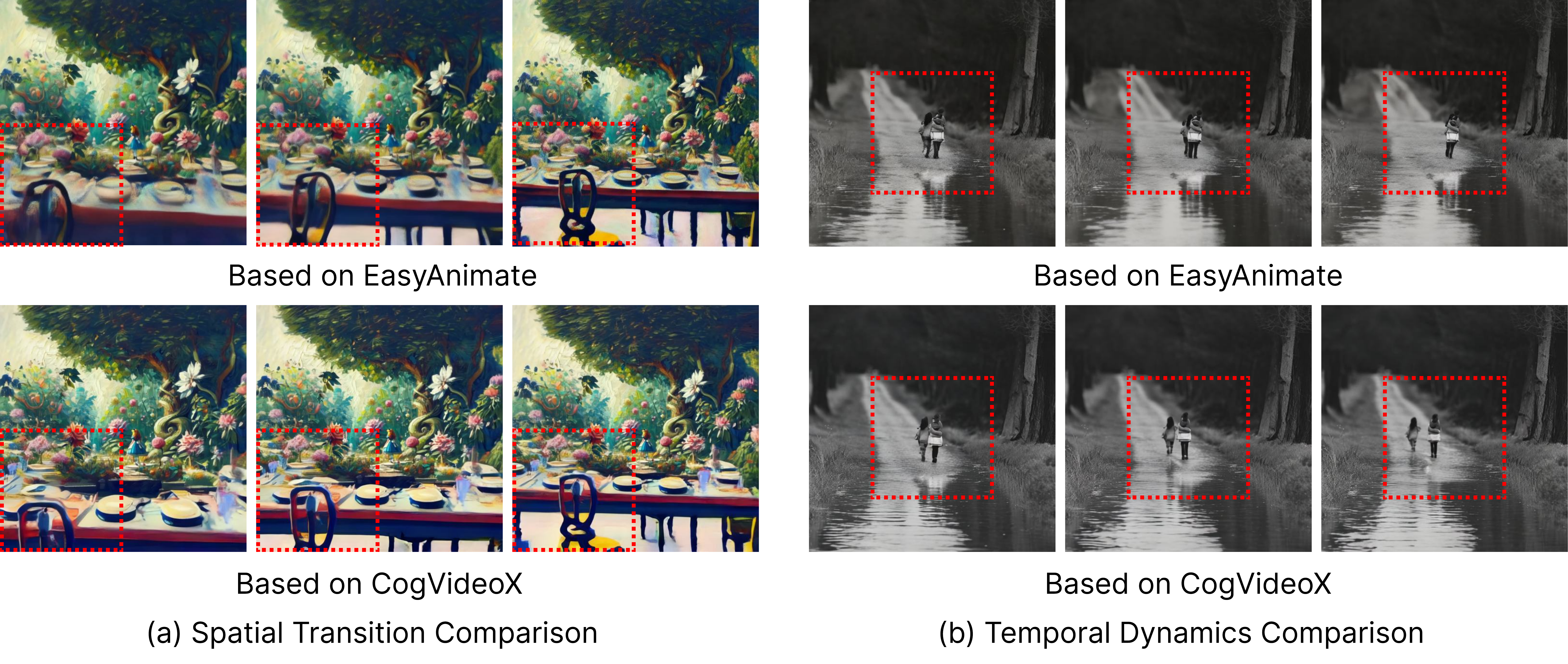} 
\caption{Improving generation quality by incorporating more advanced pre-trained models. We demonstrate the benefits of switching from EasyAnimate\cite{xu2024easyanimatehighperformancelongvideo} to CogVideoX\cite{yang2024cogvideox} (a more recent and larger model) in our framework. (a) Spatial Transition: CogVideoX generates more stable and visually appealing results for challenging view transitions in complex scenes. (b) Temporal Dynamics: CogVideoX better preserves object identity and handles occlusions in dynamic scenes with multiple interacting subjects (two walking kids), while EasyAnimate incorrectly merges the subjects.}
\label{fig: improve_with_foundation_model}
\end{figure*}

  \textbf{Framework Generality:} Our framework demonstrates good generality across different foundation models. As shown in Figure \ref{figs:different_videoModel}, two other popular video diffusion foundation models (Stable Video Diffusion \cite{Blattmann2023StableVD} and CogVideoX \cite{yang2024cogvideox}) can play the same roles as the EasyAnimate model \cite{xu2024easyanimatehighperformancelongvideo} in spatial transition generation. In Figure \ref{figs:different_MLLM}, we can observe that although prompts generated by different MLLMs have varying styles, they all effectively provide dynamic guidance for the scene. For example, the red car moves noticeably under the guidance of any of the three tested popular MLLMs, while remaining stationary under the default guidance without MLLM intervention. These results demonstrate that our framework is adaptable to different foundation models and MLLMs, which enhances its practical applicability and robustness across various implementation scenarios.

  \textbf{Comparison with general video generation model:} As shown in the Figure \ref{fig: comparison_with_general_video_model}, our method can generate stable and substantial spatial transitions (a). In contrast, existing general video generation models struggle to achieve comparable results even when camera motion directions are explicitly specified in the prompt. For example, when using the EasyAnimate model \cite{xu2024easyanimatehighperformancelongvideo} with a prompt containing the instruction "The camera zooms out steadily," the scene only undergoes minimal movement. Similarly, when using the CogVideoX 5B model \cite{yang2024cogvideox}, the scene directly performs a montage cut rather than a smooth transition.

  \textbf{Typical failures:} While our framework reaches impressive results, there is still room for improvement in terms of generation quality. We identified two main categories of typical failures: (a) Depth Prediction Failures: Current pre-trained depth estimation models often struggle with distant or small objects, leading to inaccurate 3D point cloud positioning. As illustrated in Figure \ref{fig: failure_case} (a), small birds in the distance are incorrectly reprojected as fragmented points, creating visual artifacts. These errors could potentially affect the later stages, where what should be a single bird might be misinterpreted as a flock of birds. (b) Dynamics Prediction Failure. Current pre-trained video generation models struggle within complex scenes with multiple interacting objects, producing unrealistic or inconsistent dynamics. Figure \ref{fig: failure_case} (b) demonstrates how multiple dynamic birds in the same frame may exhibit incorrect dynamics or object deformations due to these limitations. 
}

\textbf{Computation cost:} Our framework currently requires an average of 10 seconds to generate each frame of video. This considerable processing time stems from the extensive MLLM calls and video diffusion model sampling involved throughout the generation process, making real-time applications or deployment on resource-constrained devices challenging at present. In future work, more cost-effective pre-trained models for each stage could help reduce the overall computational burden.

\textbf{Scalability with Pre-trained Models:} Because our framework is modular, the generation quality of our framework can evolve with the latest advancements in pre-trained models corresponding to each module. For instance, \ref{fig: improve_with_foundation_model} demonstrates how our framework benefits from advancements in pre-trained video diffusion models. CogVideoX\cite{yang2024cogvideox} is a more recent pre-trained video diffusion model with more parameters than EasyAnimate\cite{xu2024easyanimatehighperformancelongvideo}. In (a) Spatial Transition Comparison, we observe that for challenging complex scenes with large view range transitions, switching to CogVideoX helps achieve better stability and visual quality. In (b) Temporal Dynamics Comparison, we observe that for challenging scenarios where two kids are walking together with occlusions, switching to CogVideoX helps generate more faithful dynamics (while EasyAnimate fails to handle occlusions properly and merges the two kids together).

\textbf{Ethical implications:}
The advancement of generative media technologies necessitates ongoing consideration of their ethical implications. The capabilities demonstrated by our framework could potentially be misused for creating misleading content that might spread misinformation. We emphasize the importance of responsible use and suggest implementing safeguards such as watermarking generated content, and establishing clear ethical guidelines applications.


 
\bibliography{tmm25.bib}

\bibliographystyle{IEEEtran}


 

\begin{IEEEbiography}[{\includegraphics[width=1in,height=1.25in,clip,keepaspectratio]{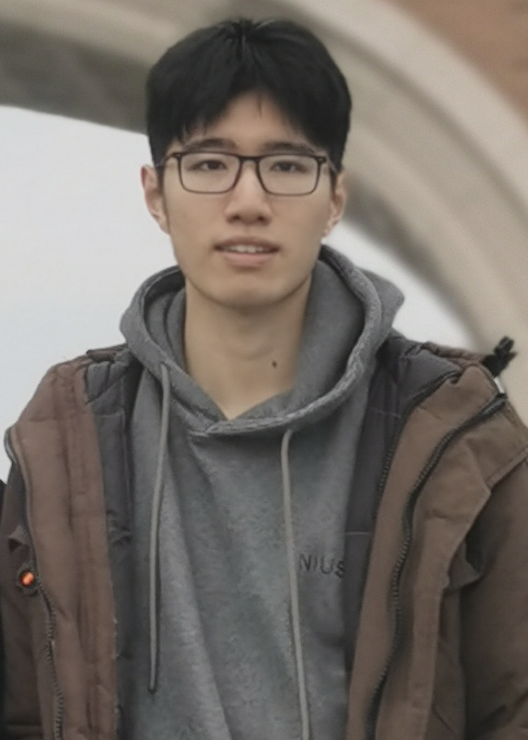}}]{Bo Pan} is currently a Ph.D. student in the State Key Lab of CAD\&CG at Zhejiang University. He received BS degree in Electrical and Computer Engineering from the University of Illinois Urbana-Champaign and Zhejiang University in 2022. His research interests include visualization and generative AI.
\end{IEEEbiography}

\begin{IEEEbiography}[{\includegraphics[width=1in,height=1.25in,clip,keepaspectratio]{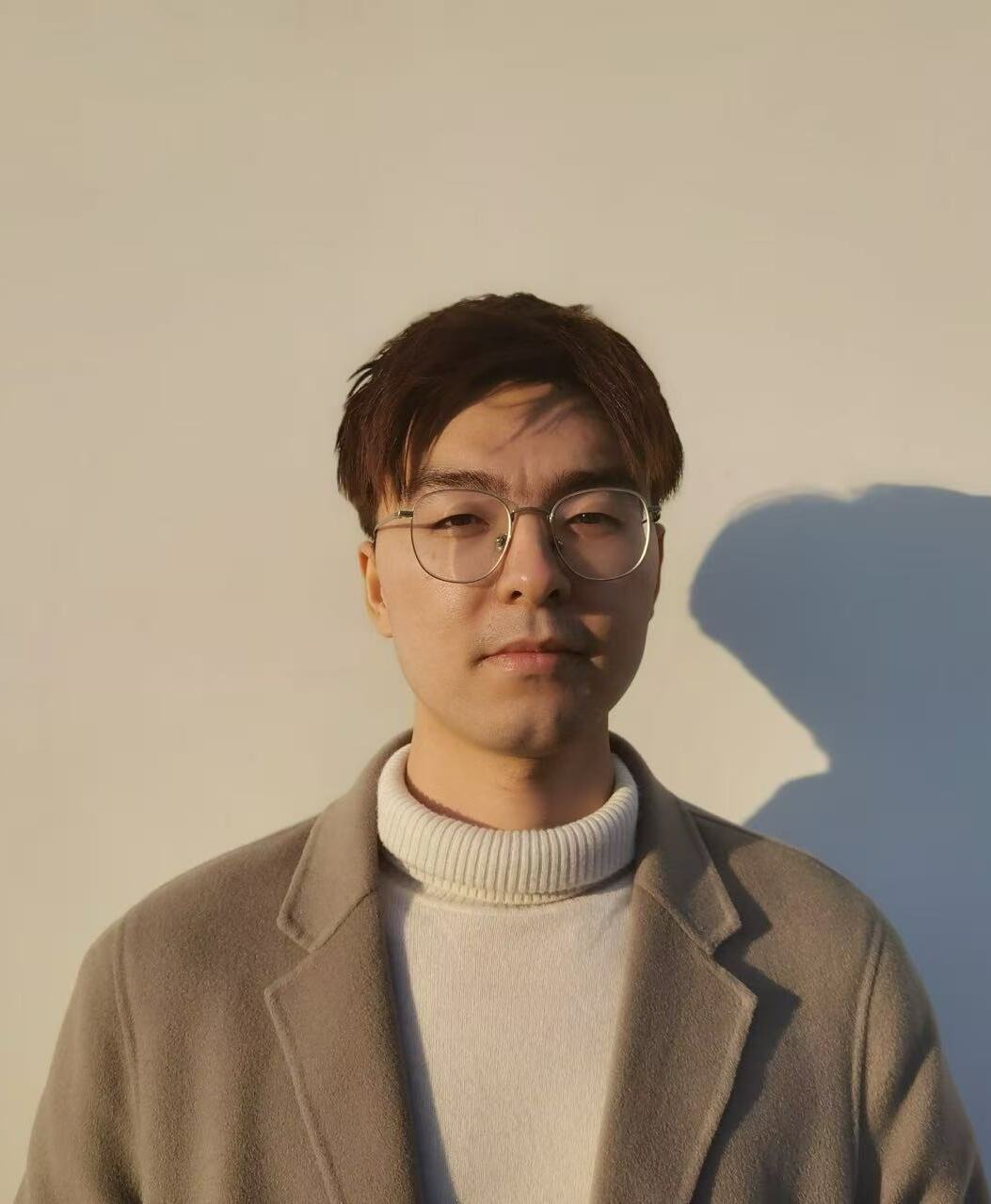}}]{Yang Chen} is currently a Researcher in HiDream.ai. He received Ph.D. degree from the University of Science and Technology of China, Hefei, China, in 2022. He was one of the core designers of the champion team in the Visual Domain Adaptation Challenge 2019 and Open World Vision Challenge 2021. He received the Second Place Best Demo Award in ACM Multimedia 2019. His research interests include generative AI and 3D vision. 
\end{IEEEbiography}

\begin{IEEEbiography}[{\includegraphics[width=1in,height=1.25in,clip,keepaspectratio]{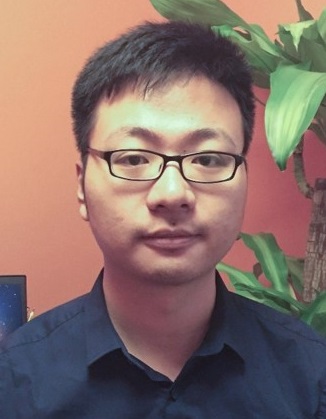}}]{Yingwei Pan}
is currently a Technical Director in HiDream.ai. His research interests include vision and language, and visual content understanding. He is the principal designer of the top-performing multimedia analytic systems in international competitions such as COCO Image Captioning, Visual Domain Adaptation Challenge 2018, ActivityNet Dense-Captioning Events in Videos Challenge 2017, and MSR-Bing Image Retrieval Challenge 2014 and 2013. He received Ph.D. degree in Electrical Engineering from University of Science and Technology of China in 2018.
\end{IEEEbiography}

\begin{IEEEbiography}[{\includegraphics[width=1in,height=1.25in,clip,keepaspectratio]{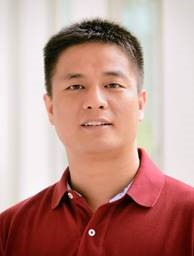}}]{Ting Yao}
is currently the CTO of HiDream.ai, a high-tech startup company focusing on generative intelligence for creativity. Previously, he was a Principal Researcher with JD AI Research in Beijing, China and a Researcher with Microsoft Research Asia in Beijing, China. Dr. Yao has co-authored more than 100 peer-reviewed papers in top-notch conferences/journals. He has developed one standard 3D Convolutional Neural Network, i.e., Pseudo-3D Residual Net, for video understanding, and his video-to-text dataset of MSR-VTT has been used by 400+ institutes worldwide. He serves as an associate editor of IEEE Transactions on Multimedia, Pattern Recognition Letters, and Multimedia Systems. His works have led to many awards, including 2015 ACM-SIGMM Outstanding Ph.D. Thesis Award, 2019 ACM-SIGMM Rising Star Award, 2019 IEEE-TCMC Rising Star Award, 2022 IEEE ICME Multimedia Star Innovator Award, and the winning of 10+ championship in worldwide competitions.
\end{IEEEbiography}

\begin{IEEEbiography}[{\includegraphics[width=1in,height=1.25in,clip,keepaspectratio]{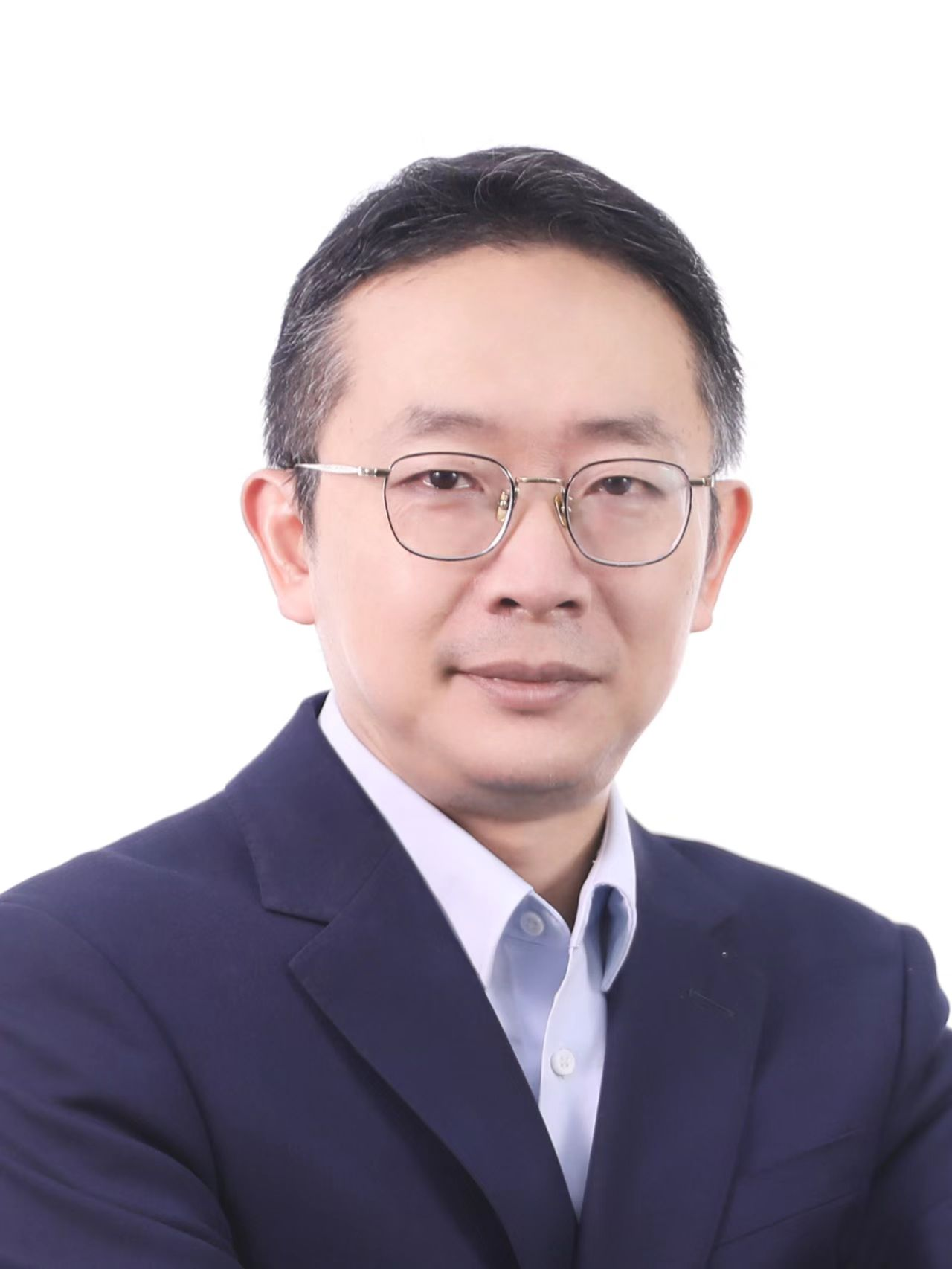}}]{Wei Chen}
is a professor in the State Key Lab of CAD\&CG at Zhejiang University. His current research interests include visualization and visual analytics. He has published more than 80 IEEE/ACM Transactions and IEEE VIS papers. He actively served in many leading conferences and journals, like IEEE PacificVIS steering committee, ChinaVIS steering committee, paper cochairs of IEEE VIS, IEEE PacificVIS, IEEE LDAV and ACM SIGGRAPH Asia VisSym. He is an associate editor of IEEE TVCG, IEEE TBG, ACM TIST, IEEE T-SMC-S, IEEE TIV, IEEE CG\&A, FCS, and JOV. More information can be found at: \url{http://www.cad.zju.edu.cn/home/chenwei}
\end{IEEEbiography}

\begin{IEEEbiography}[{\includegraphics[width=1in,height=1.25in,clip,keepaspectratio]{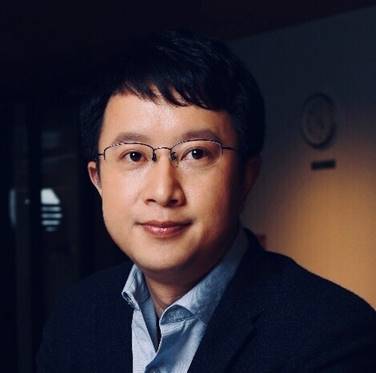}}]{Tao Mei}
(Fellow, IEEE) is the Founder and CEO of HiDream.ai. Previously, He was a Vice President of JD.COM and a Senior Research Manager of Microsoft Research. He has authored or co-authored over 200 publications (with 12 best paper awards) in journals and conferences, 10 book chapters, and edited five books. He holds over 25 US and international patents. He is or has been an Editorial Board Member of IEEE Trans. on Image Processing, IEEE Trans. on Circuits and Systems for Video Technology, IEEE Trans. on Multimedia, ACM Trans. on Multimedia Computing, Communications, and Applications, Pattern Recognition, etc. He is the General Co-chair of IEEE ICME 2019, the Program Co-chair of ACM Multimedia 2018, IEEE ICME 2015 and IEEE MMSP 2015.
	
Tao received B.E. and Ph.D. degrees from the University of Science and Technology of China, Hefei, China, in 2001 and 2006, respectively. He is a Fellow of IEEE (2019), a Fellow of IAPR (2016), a Distinguished Scientist of ACM (2016), and a Distinguished Industry Speaker of IEEE Signal Processing Society (2017).
\end{IEEEbiography}




\vfill

\end{document}